\documentclass[letterpaper]{article} 
\usepackage{aaai2026}  
\usepackage{times}  
\usepackage{helvet}  
\usepackage{courier}  
\usepackage[hyphens]{url}  
\usepackage{graphicx} 
\usepackage{natbib}  
\usepackage{caption} 
\usepackage{adjustbox}
\usepackage{algorithm}
\usepackage{algorithmic}
\usepackage{amsmath}
\usepackage{amssymb}
\usepackage{booktabs}
\usepackage{multirow}
\usepackage{pifont}
\usepackage{subcaption}

\newcommand{\figref}[1]{Fig.~\ref{fig:#1}}
\newcommand{\tabref}[1]{Table~\ref{tab:#1}}

\nocopyright

\usepackage{newfloat}
\usepackage{listings}
\DeclareCaptionStyle{ruled}{labelfont=normalfont,labelsep=colon,strut=off} 
\floatstyle{ruled}
\newfloat{listing}{tb}{lst}{}
\floatname{listing}{Listing}
\title{Group Orthogonal Low-Rank Adaptation for RGB-T Tracking}
\author{
    Zekai Shao,
    Yufan Hu,
    Jingyuan Liu,
    Bin Fan,
    Hongmin Liu\thanks{Corresponding author.}
}
\affiliations{
    School of Intelligence Science and Technology, University of Science and Technology Beijing\\

    melan.shao@foxmail.com, huyufanqaixuan@gmail.com, ljy.ustb@foxmail.com, \\ bin.fan@ieee.org, hmliu\_82@163.com
}

\begin{document}

\maketitle

\begin{abstract}
Parameter-efficient fine-tuning has emerged as a promising paradigm in RGB-T tracking, enabling downstream task adaptation by freezing pretrained parameters and fine-tuning only a small set of parameters. This set forms a rank space made up of multiple individual ranks, whose expressiveness directly shapes the model's adaptability. However, quantitative analysis reveals low-rank adaptation exhibits significant redundancy in the rank space, with many ranks contributing almost no practical information. This hinders the model's ability to learn more diverse knowledge to address the various challenges in RGB-T tracking. To address this issue, we propose the Group Orthogonal Low-Rank Adaptation (GOLA) framework for RGB-T tracking, which effectively leverages the rank space through structured parameter learning. Specifically, we adopt a rank decomposition partitioning strategy utilizing singular value decomposition to quantify rank importance, freeze crucial ranks to preserve the pretrained priors, and cluster the redundant ranks into groups to prepare for subsequent orthogonal constraints. We further design an inter-group orthogonal constraint strategy. This constraint enforces orthogonality between rank groups, compelling them to learn complementary features that target diverse challenges, thereby alleviating information redundancy. Experimental results demonstrate that GOLA effectively reduces parameter redundancy and enhances feature representation capabilities, significantly outperforming state-of-the-art methods across four benchmark datasets and validating its effectiveness in RGB-T tracking tasks.
\end{abstract}

\begin{links}
    \link{Code}{https://github.com/MelanTech/GOLA}
\end{links}

\begin{figure}[!t]
    \centering
    \includegraphics[width=\linewidth]{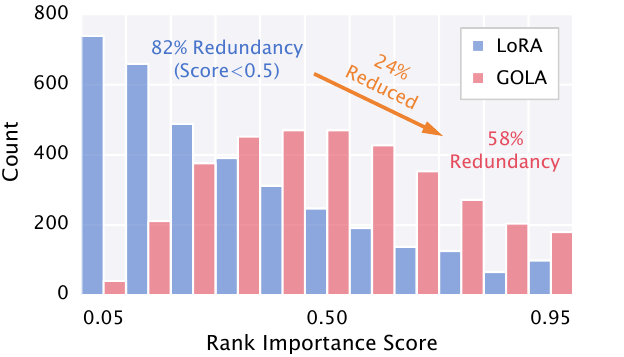}
    \caption{Comparison of rank importance score distribution between LoRA and our proposed GOLA. The rank space of LoRA exhibits significant redundancy, while our proposed GOLA effectively mitigates this issue.}
    \label{fig:motivation}
\end{figure}

\section{Introduction}

RGB-T tracking combines the complementary strengths of visible and infrared data, significantly enhancing the robustness in challenging scenarios, such as low-light or occluded conditions. It has been applied in multiple domains, demonstrating its effectiveness in complex scenarios.

Recently, most RGB-T trackers adopt a prompt-tuning paradigm, integrating lightweight prompt modules \cite{vipt, bat, dmd} to facilitate multimodal fusion while maintaining an acceptable speed. However, this prompt-tuning approach typically relies on feature dimensionality reduction to balance efficiency and performance, which limits the capacity of interaction between modalities to express information in a low-rank space and weakens the model's overall performance. LoRA \cite{lora} applies the low-rank concept to the parameter level by freezing the backbone and optimizing only the low-rank matrices, thereby avoiding information loss and enabling parameter merging during the inference stage, which enhances inference efficiency. Nonetheless, in complex RGB-T tracking tasks, the redundant ranks of LoRA do not effectively facilitate the acquisition of diverse knowledge, which directly undermines its ability to address various challenges. As shown in \figref{motivation}, after training the model with LoRA, we decompose all ranks in its parameters and plot the histogram of singular values. The results indicate significant redundancy in the rank space of LoRA, with only a few ranks becoming dominant. This phenomenon arises from LoRA's compression of all information into a low-rank space during downstream adaptation, making it difficult to address the diverse challenges present in RGB-T tracking. In training, to achieve rapid convergence to an optimal solution, it tends to prioritize the integration of a few crucial ranks that encapsulate the most significant task, which overshadows the pretrained priors. Simultaneously, the remaining ranks remain unactivated due to a lack of targeted optimization signals, failing to capture the fine-grained features necessary for specific challenges, while also struggling to supplement the expressive blind spots uncovered by crucial ranks. Ultimately, this severely limits the overall expressiveness of the rank space, hampering the model's ability to handle diverse challenges in RGB-T tracking.

To address the issues above, we propose a novel Group Orthogonal Low-Rank Adaptation (GOLA) framework for RGB-T tracking, aiming to preserve the pretrained priors while fully activating the expressive potential of the redundant rank space. First, we introduce a rank decomposition partitioning strategy that quantifies the significance of each rank by performing singular value decomposition on the LoRA parameter matrix and categorizes these ranks into crucial and redundant ones. We freeze the crucial ranks to retain the strong generalization capability of the pretrained weights. This operation is conducted offline and does not incur additional training burdens. Furthermore, we utilize clustering to partition the redundant rank components into parallel rank subsets, establishing a foundation for subsequent orthogonal constraints. We then introduce an inter-group orthogonal constraint strategy by constructing an orthogonal loss, which enforces the parameters of two randomly sampled groups to maintain an orthogonal relationship. This strategy ensures that the parameters of each group learn mutually independent and complementary feature transformations, effectively avoiding overlap and redundancy in feature learning. Ultimately, the frozen critical ranks robustly preserve the existing priors, while the grouped redundant ranks focus on learning knowledge tailored to different challenges under orthogonal constraints. This synergistic effect collectively enhances the model's tracking performance in complex RGB-T scenarios. In summary, our contributions are threefold:

\begin{itemize}
    \item We propose the Group Orthogonal Low-Rank Adaptation (GOLA) framework for RGB-T tracking, addressing the shortcomings of rank space redundancy and limited expressive capability, enhancing the model's ability to handle diverse challenges.
    \item In our framework, we design a rank decomposition strategy to freeze the crucial ranks and group the redundant ranks, along with an inter-group orthogonal constraint that enforces complementary feature learning across groups to enhance the model's representation ability.
    \item Experimental results validate the effectiveness of the proposed method in RGB-T tracking, demonstrating that our method achieves optimal performance and efficiency.
\end{itemize}

\begin{figure*}[!t]
    \centering
    \includegraphics[width=\linewidth]{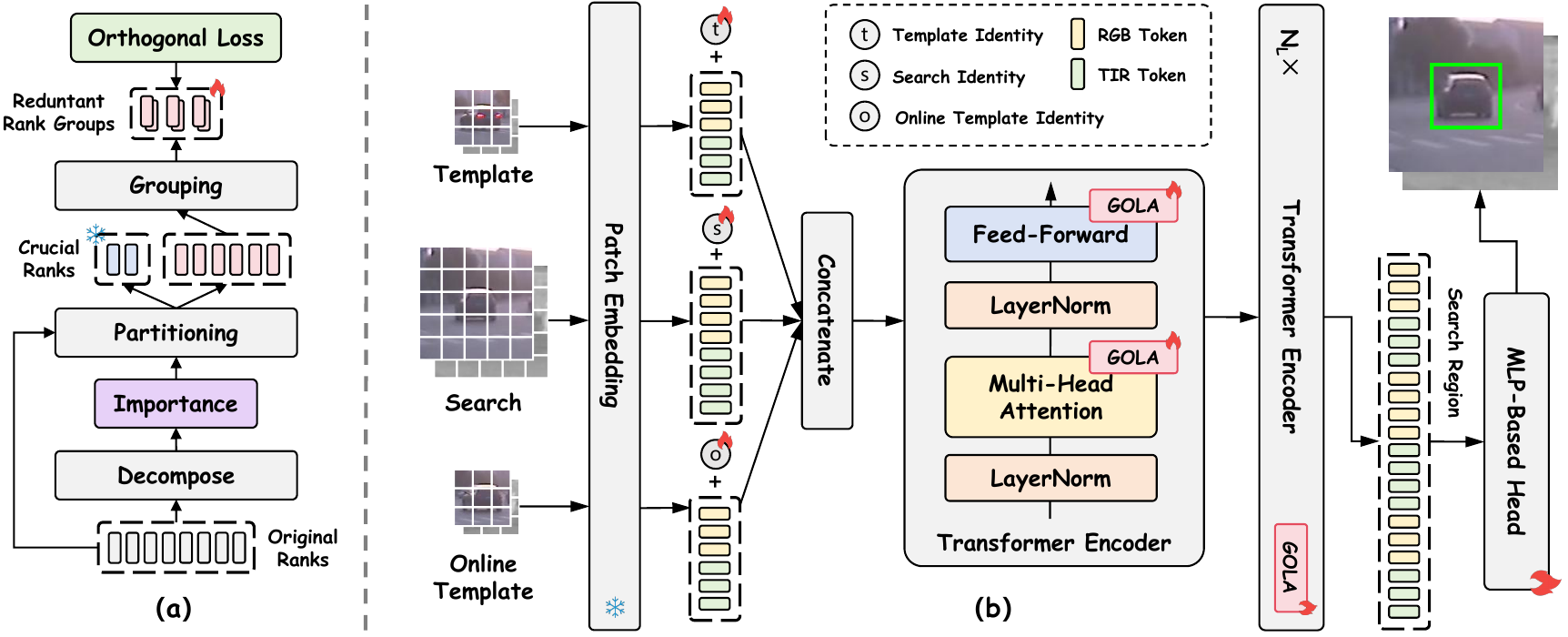}
    \caption{(a) Our proposed Group Orthogonal Low-Rank Adaptation (GOLA) framework. We decompose pretrained ranks into crucial ranks and redundant rank groups, freezing the crucial ranks to retain generalization. Our inter-group orthogonal constraint boosts the expressiveness of the redundant rank groups, fully utilizing the redundant rank space. (b) The overall architecture of our tracking framework. GOLA is applied to each linear layer of the backbone for fine-tuning.}
    \label{fig:pipeline}
\end{figure*}

\section{Related Works}

\subsection{RGB-T Tracking}

With the aid of visible and infrared modalities, RGB-T tracking significantly enhances tracking robustness in complex scenes through information fusion. Recent RGB-T trackers are mainly categorized into full fine-tuning, prompt-tuning, online adaptation, and low-rank adaptation methods based on their training strategies. Full fine-tuning approaches fine-tune trackers pretrained on the RGB modality to adapt to the RGB-T task. TBSI \cite{tbsi} uses a dual-stream backbone, treats templates as a medium to bridge RGB and infrared modalities, and incorporates multimodal information via template updates. USTrack \cite{ustrack} first unifies features into a single backbone, solving issues of separate modal feature extraction and slow speed. Full fine-tuning achieves good tracking performance but has long training cycles and high resource consumption. Moreover, scarce RGB-T training data makes existing methods prone to overfitting and performance degradation during full fine-tuning. Prompt-tuning-based methods \cite{vipt, tatrack, cfbt} have reduced training overhead by fine-tuning lightweight prompt modules, while maintaining tracking performance comparable to that of full fine-tuning methods. Nevertheless, prompt-tuning-based methods rely on additional modules and parameters, severely limiting model running efficiency. PURA \cite{pura} proposes a framework based on parameter decomposition and recovery to achieve adaptation for RGB-T tracking in domain shift scenarios. UnTrack \cite{untrack} introduces LoRA \cite{lora} for low-rank fine-tuning of the backbone to reduce the number of trainable parameters. However, it still employs prompt-tuning methods for the infrared modality branch, which similarly relies on additional parameters. Moreover, the low-rank characteristics of LoRA hinder its ability to learn diverse knowledge to tackle different challenges. In this work, we focus on reducing parameter redundancy through the structured management of rank spaces and orthogonal constraints, thereby achieving efficient cross-modal transfer.

\subsection{Parameter-Efficient Fine-Tuning}

As the core paradigm for lightweight adaptation of pretrained models, Parameter-Efficient Fine-Tuning (PEFT) has recently demonstrated successful extensions to computer vision, providing solutions for model training in resource-constrained scenarios. PEFT can be broadly categorized into adapter-based methods, prompt-based methods, and low-rank adaptation (LoRA) \cite{lora}. Adapter-based methods \cite{adapter, convpass} enable transfer learning by introducing lightweight modules into pretrained models. Prompt-based methods \cite{vpt, prefix-tuning, prompt-tuning} focus on utilizing carefully designed prompts to guide pretrained models in generating responses for specific tasks. LoRA achieves fine-tuning by decomposing weight updates into low-rank matrices during the adaptation process, leveraging the inherent low-rank structure of weight updates to enable adaptation with low computational overhead. To enhance learning efficiency, some variants \cite{adalora, dora} dynamically adjust parameters through weight decomposition, optimizing the learning process. Recently, methods based on a mixture of experts have shown significant effectiveness in multitask scenarios. MOLA \cite{mola} markedly improves fine-tuning outcomes by setting varying numbers of experts at different layers. MixLoRA \cite{mixlora} employs multiple LoRA-based experts while introducing load-balancing losses to address the imbalance issues among routers. Furthermore, LoRA has revealed significant potential in tracking tasks, with LoRAT \cite{lorat} and UnTrack \cite{untrack} utilizing this technique for the fine-tuning of the backbone, thereby enhancing tracking performance while reducing the computational burden.

\section{Methodology}

This paper presents the Group Orthogonal Low-Rank Adaptation (GOLA) framework for RGB-T tracking, which enhances tracking performance by rank partition and orthogonal constraints on pretrained ranks. This section outlines the tracker architecture and provides details on the method's implementation. The overall framework is depicted in \figref{pipeline}.

\subsection{Tracking Framework}

As shown in \figref{pipeline}, our method adopts a simple single-stream tracking framework. Let $\mathbf{I}_v^z, \mathbf{I}_t^z$ be template images, $\mathbf{I}_v^x, \mathbf{I}_t^x$ be search images, and $\mathbf{I}_v^o, \mathbf{I}_t^o$ be online template images. Here, $v$ and $t$ represent the visible and thermal modalities, respectively. We first transform these images into token sequences, resulting in $\mathbf{Z}_v, \mathbf{Z}_t \in \mathbb{R}^{n_z \times c}$, $\mathbf{X}_v, \mathbf{X}_t \in \mathbb{R}^{n_x \times c}$, and $\mathbf{O}_v, \mathbf{O}_t \in \mathbb{R}^{n_z \times c}$. Here, $\mathbf{Z}$, $\mathbf{X}$, and $\mathbf{O}$ represent the template, the search region, and the online template, respectively. $n_z$ and $n_x$ represent the number of tokens for the template and the search region, respectively, and $c$ signifies the number of feature channels. In addition, we provide unique trainable type embeddings \cite{lorat} for each category of tokens to prevent information confusion.

The tokens are then concatenated to form the final embedding vector denoted as $h = [\mathbf{Z}_v; \mathbf{Z}_t; \mathbf{X}_v; \mathbf{X}_t; \mathbf{O}_v; \mathbf{O}_t]$. Subsequently, we process $h$ through a series of encoder layers for joint feature extraction. Our GOLA uses the same fine-tuning strategy as LoRA \cite{lora}. For each linear layer parameter $\mathbf{W}$ in the backbone, we introduce two low-rank matrices $\mathbf{A}$ and $\mathbf{B}$ to update it as follows:
\begin{equation}
    h' = \mathbf{W}h + \mathbf{B}\mathbf{A}h,
\end{equation}

\noindent where $\mathbf{A} = \{ \mathbf{a}_1, \mathbf{a}_2, \dots , \mathbf{a}_r \} \in \mathbb{R}^{r \times c}$ and $\mathbf{B} = \{ \mathbf{b}_1, \mathbf{b}_2, \dots , \mathbf{b}_r \} \in \mathbb{R}^{c \times r}$, and $r \ll c$. In the inference stage, we use the following formula for parameter merging: 
\begin{equation}
    \mathbf{W}' = \mathbf{W} + \mathbf{B}\mathbf{A}.
\end{equation}
Subsequently, the search region features produced by the backbone are fed into the prediction head to obtain the final results. Additionally, we use the max score map value from the prediction head as confidence, comparing it with the threshold $\tau$ to update the online template.

\subsection{Group Orthogonal Low-Rank Adaptation}

\noindent \textbf{Rank Decomposition Partition.} Our rank decomposition partition aims to leverage singular value decomposition to quantify the importance of ranks, thereby distinguishing between critical and redundant ranks and grouping redundant ranks for structured learning. Notably, this process is offline and does not affect the training workflow. The matrix $\mathbf{B}$ exhibits a significant rank correlation with task specificity compared to the matrix $\mathbf{A}$, which functions as a general feature extractor \cite{hydralora}. This stronger association of $\mathbf{B}$ with modality-specific tasks leads to its use as the basis for partitioning. Specifically, we first perform singular value decomposition on matrix $\mathbf{B}$ to obtain the rank importance:
\begin{equation}
    \mathbf{\Sigma}, \mathbf{V} \leftarrow \textbf{SVD} \left(\bar{\mathbf{B}} \right),
\end{equation}

\noindent where $\bar{\mathbf{B}}$ represents the mean-centered matrix of $\mathbf{B}$, singular values $\mathbf{\Sigma} \in \mathbb{R}^{r}$ represent the importance of each rank, and the singular vectors $\mathbf{V} \in \mathbb{R}^{c \times c}$ represent the principal direction of each rank, with both ordered in descending order of importance. Subsequently, we select the top-$k$ singular vectors $V_k \in \mathbb{R}^{k \times c}$ as reference vectors and use the top-$k$ singular values $\Sigma_k \in \mathbb{R}^{k}$ as weights. We compute the importance score $\mathbf{S}$ for each rank by calculating the weighted similarity between each rank and the reference ranks. This process is represented as:
\begin{equation}
    \mathbf{S} = \Vert \bar{\mathbf{B}}^{\top} \mathbf{V}_k^{\top} \odot \mathbf{\Sigma}_k \Vert_{2},
\end{equation}

\noindent where $\odot$ represents the hadamard product, and $\Vert \cdot \Vert_2$ denotes $L_2$ column normalization. After obtaining the importance score $\mathbf{S} \in \mathbb{R}^{r}$, we sort all ranks in descending order based on $\mathbf{S}$ to determine their priority. Let the sorted index be denoted as $\sigma$, which satisfies:
\begin{equation}
    \mathbf{S}_{\sigma_1} \geq \mathbf{S}_{\sigma_2} \geq \cdots \geq \mathbf{S}_{\sigma_r},
\end{equation}

\noindent where $\sigma_i$ denotes the original index of the $i$-th sorted element after sorting. Subsequently, we perform the same rank partitioning on $\mathbf{A}$ and $\mathbf{B}$ to obtain crucial ranks $\mathbf{A}_c$ and $\mathbf{B}_c$:
\begin{equation}
    \begin{split}
        \mathbf{A}_c &=  \{ \mathbf{a}_{\sigma_1}, \mathbf{a}_{\sigma_2}, \dots , \mathbf{a}_{\sigma_k} \}, \\
        \mathbf{B}_c &=  \{ \mathbf{b}_{\sigma_1}, \mathbf{b}_{\sigma_2}, \dots , \mathbf{b}_{\sigma_k} \},
    \end{split}
\end{equation}

\noindent where $\mathbf{a}_{\sigma_i}$ and $\mathbf{b}_{\sigma_i}$ represent the selections of ranks using the $i$-th sorted index. It is noteworthy that performing the same sorting operation on $\mathbf{A}$ and $\mathbf{B}$ produces new weights that are equivalent to the original weights, thereby not affecting the training process. Since $\mathbf{A}_c$ and $\mathbf{B}_c$ serve as the principal components for essential feature transformation, we freeze them to preserve the primary functions of the pretrained weights. In contrast, the remaining ranks, being of lesser importance, are considered as redundant ranks $\mathbf{A}_u$ and $\mathbf{B}_u$. We fine-tune them to learn the knowledge brought by the new modality. To ensure that these redundant ranks can discover a more diverse set of expertise, we divide them into $n$ groups for subsequent orthogonal constraints. Specifically, we continue to use $\mathbf{B}$ as the reference and perform clustering on the redundant ranks to obtain the group indices:
\begin{equation}
    \{ \mathbf{G}_1, \mathbf{G}_2, \dots, \mathbf{G}_n \} = \mathbf{\Gamma}(\mathbf{B}_u, n),
\end{equation}

\noindent where $\mathbf{G}_i$ denotes the index set of the $i$-th group, with $\left| \mathbf{G}_i \right| = (r-k)/n$. $\mathbf{\Gamma}(\cdot)$ represents the clustering operation. We employ constrained k-means \cite{c-kmeans} to balance the rank capacity of each group. Then, we use $\mathbf{G}_i$ to group the redundant ranks:
\begin{equation}
    \begin{split}
        \mathbf{A}_{u_i} = \{ \mathbf{a}_j \mid j \in \mathbf{G}_i \}, \quad i = 1, 2, \dots, n, \\
        \mathbf{B}_{u_i} = \{ \mathbf{b}_j \mid j \in \mathbf{G}_i \}, \quad i = 1, 2, \dots, n. \\
    \end{split}
\end{equation}

\noindent After the completion of the redundant rank partitioning, the subsequent orthogonal constraints enforce that each group occupies non-overlapping feature spaces to achieve inter-group complementarity.

\noindent\textbf{Inter-Group Orthogonal Constraint.} Our inter-group orthogonal constraints aim to enhance the independence of feature learning among different rank groups, thereby reducing the parameter redundancy in rank spaces. Specifically, $\mathbf{A}$ functions as a universal feature extractor, capturing the fundamental standard features within bimodal data \cite{hydralora}. To enhance both the diversity and discriminative power of these general features, a channel orthogonality constraint is imposed on $\mathbf{A}$. In contrast, $\mathbf{B}$ focuses on learning task-specific knowledge, so rank orthogonal constraints are applied to ensure that the task knowledge carried by different ranks is complementary. The orthogonal constraint loss can be expressed as:
\begin{equation}
    \mathcal{L}_{orth} = \sum_{i \neq j} \left( \left| \mathbf{A}_{u_i}^\top \mathbf{A}_{u_j} \right| + \left| \mathbf{B}_{u_i}^\top \mathbf{B}_{u_j} \right| \right).
\end{equation}

\noindent Notably, calculating the above loss for all pairs of rank groups incurs a substantial computational burden. Therefore, we randomly select only one pair of different rank groups to compute the orthogonality loss at each iteration, thereby reducing the computational load.

\subsection{Objective Loss} For the tracking task, we use binary cross-entropy loss for classification and GIoU loss \cite{giou} for regression. The overall loss function is summarized as follows:
\begin{equation}
    \mathcal{L}=\mathcal{L}_{cls}+\mathcal{L}_{reg}+ \lambda \cdot \mathcal{L}_{orth},
\end{equation}

\noindent where $\mathcal{L}_{cls}$ represents the binary cross-entropy loss, while $\mathcal{L}_{reg}$ represents the GIoU loss. $\lambda$ is the balance factor.

\begin{table*}[!t]
    \centering
    \small
    \begin{tabular}{l|ccccccccc|c}
    \toprule
        \multirow{2}*{Methods} & \makebox[0pt][l]{\hspace{4pt}GTOT} & & \makebox[0pt][l]{\hspace{-5pt}RGBT210} & & \makebox[0pt][l]{\hspace{-3pt}RGBT234} & & \makebox[0pt][l]{\hspace{14pt}LasHeR} & & & Speed \\
        \cline{2-10}
            & MPR & MSR & PR & SR & MPR & MSR & PR & NPR & SR & (\textit{fps}) \\
        \midrule
            APFNet~\cite{apfnet} & 90.5 & 73.7 & 79.9 & 54.9 & 82.7 & 57.9 & 50.0 & 43.9 & 36.2 & 1 \\
            CAT++~\cite{cat++} & 91.5 & 73.3 & 82.2 & 56.1 & 84.0 & 59.2 & 50.9 & 44.4 & 35.6 & 14\\
            ViPT~\cite{vipt} & - & - & - & - & 83.5 & 61.7 & 65.1 & - & 52.5 & - \\
            TBSI~\cite{tbsi} & - & - & 85.3 & 62.5 & 87.1 & 63.7 & 69.2 & 65.7 & 55.6 & 36 \\
            TATrack~\cite{tatrack} & - & - & 85.3 & 61.8 & 87.2 & 64.4 & 70.2 & 66.7 & 56.1 & 26 \\
            STMT~\cite{stmt} & - & - & 83.0 & 59.5 & 86.5 & 63.8 & 67.4 & 63.4 & 53.7 & 39 \\
            BAT~\cite{bat} & - & - & - & - & 86.8 & 64.1 & 70.2 & - & 56.3 & - \\
            CKD~\cite{ckd} & \textbf{93.2} & \underline{77.2} & \underline{88.4} & \underline{65.2} & 90.0 & 67.4 & 73.2 & 69.3 & 58.1 & \underline{96} \\
            SDSTrack~\cite{sdstrack} & - & - & - & - & 84.8 & 62.5 & 66.5 & - & 53.1 & 21 \\
            CFBT~\cite{cfbt} & - & - & 87.7 & 63.0 & 89.9 & 65.9 & 73.2 & \underline{69.5} & 58.4 & - \\
            DMD~\cite{dmd} & 92.4 & 76.8 & 87.0 & 63.7 & 89.3 & 66.7 & 72.6 & 68.6 & 57.6 & 17 \\
            STTrack~\cite{sttrack} & - & - & - & - & 89.8 & 66.7 & \underline{76.0} & - & 60.3 & 36 \\
            SeqTrackv2-B384~\cite{seqtrackv2} & - & - & - & - & 90.0 & 66.3 & 71.5 & - & 56.2 & 15 \\
            SUTrack-B384~\cite{sutrack} & - & - & - & - & \underline{92.1} & \underline{69.2} & 75.8 & - & \underline{60.9} & 32 \\
        \midrule
            GOLA-B & \underline{92.8} & \textbf{78.5} & \textbf{90.9} & \textbf{67.0} & \textbf{92.2} & \textbf{69.5} & \textbf{77.5} & \textbf{73.9} & \textbf{61.6} & \textbf{125} \\
        \midrule[\heavyrulewidth]
            SeqTrackv2-L384~\cite{seqtrackv2} & - & - & - & - & 91.3 & 68.0 & 76.7 & - & 61.0 & 5 \\
            SUTrack-L384~\cite{sutrack} & - & - & - & - & \textbf{93.7} & \underline{70.3} & 76.9 & - & \textbf{61.9} & \underline{12} \\
        \midrule
            GOLA-L & \textbf{95.3} & \textbf{80.9} & \textbf{92.0} & \textbf{68.7} & \underline{92.8} & \textbf{71.3} & \textbf{78.1} & \textbf{74.5} & \textbf{61.9} & \textbf{64} \\
        \bottomrule
    \end{tabular}
    \caption{Comparison of state-of-the-art performance across the GTOT, RGBT210, RGBT234 and LasHeR datasets. \textbf{Bold} indicates the highest performance, and \underline{underlined} indicates the second-highest performance.}
    \label{tab:evaluation}
\end{table*}

\begin{table}[!t]
    \centering
    \small
    \setlength{\tabcolsep}{1mm}
    \begin{tabular}{c|ccccccc}
        \toprule
        \multirow{2}*{Variant} & Template & Search & Params & Trainable & FLOPs \\
        & Resolution& Resolution & (M) & (\%) & (G) \\
        \midrule
        GOLA-B & 112$\times$112 & 224$\times$224 & 99 & 10 & 85 \\
        GOLA-L & 112$\times$112 & 224$\times$224 & 336 & 8 & 284 \\
        \bottomrule
    \end{tabular}
    \caption{Details of input resolution, parameter count, and efficiency for different variants.}
    \label{tab:variant}
\end{table}

\section{Experiments}

\subsection{Implementation Details}

We utilize LoRAT \cite{lorat} as a baseline and initialize the model with its pretrained weights. As shown in \tabref{variant}, we provide two different scale variants, GOLA-B and GOLA-L. In all variants, the model trains for 10 epochs on four NVIDIA A40 GPUs, with a batch size of 128, processing 131,072 image pairs per epoch. The rank $r$ is set to 64, the number of crucial ranks $k$ is set to 16, and the number of redundant rank groups $n$ is set to 8. The balance factor $\lambda$ is set to $1.4 \times 10^{-3}$. The threshold $\tau$ is set to 0.84. During the inference phase, we test the inference speed using a single NVIDIA RTX 3090 GPU.

\subsection{Comparison with State-of-the-Arts}

\noindent \textbf{Evaluation on GTOT \cite{gtot}.} GTOT includes 50 precisely paired video sequences, each with an infrared and a grayscale video. As shown in \tabref{evaluation}, GOLA-L establishes a new state-of-the-art performance with 95.3\% MPR and 80.9\% MSR. This result surpasses all previous methods and demonstrates a significant improvement over the strong baseline of CKD. GOLA-B also yields robust results, validating that our architecture exhibits strong generalizability even under the basic configuration.

\noindent \textbf{Evaluation on RGBT210 \cite{rgbt210}.} RGBT210 is a precisely annotated dataset of 210 visible and infrared video pairs, comprising over 210,000 frames. As shown in \tabref{evaluation}, GOLA demonstrates significant advantages on RGBT210. GOLA-B achieves a PR of 90.9\% and a SR of 67.0\%, showing notable improvements compared to methods such as CKD and CFBT. Relative to GOLA-B, GOLA-L improves the PR and SR metrics to 92.0\% and 68.7\%, respectively, resulting in a further performance gain.

\noindent \textbf{Evaluation on RGBT234 \cite{rgbt234}.} RGBT234 contains 234 precisely aligned visible-infrared video sequence pairs, with roughly 234,000 frames in total. As shown in \tabref{evaluation}, GOLA demonstrates a significant performance advantage on RGBT234. GOLA-B performs comparably to SUTrack-L384. Furthermore, GOLA-L enhances the MPR to 92.8\% and the MSR to 71.3\%, with the MSR showing an improvement of 1.0\% compared to SUTrack-L384. The experimental results validate the effectiveness of our approach.

\noindent \textbf{Evaluation on LasHeR \cite{lasher}.} LasHeR comprises 1,224 visible-infrared video pairs with over 734,800 frame pairs in total. As shown in \tabref{evaluation}, GOLA demonstrates remarkable performance, surpassing competitive approaches such as SeqTrackv2 and SUTrack. GOLA-B achieves 77.5\% of PR, 73.9\% of NPR, and 61.6\% of SR, outperforming larger trackers such as SeqTrackv2-L384 and SUTrack-L384. GOLA-L further elevates the PR to 78.1\%, showing improvements of 1.4\% and 1.2\% compared to SeqTrackv2-L384 and SUTrack-L384, respectively.

\begin{figure}[!t]
    \centering
    \includegraphics[width=\linewidth]{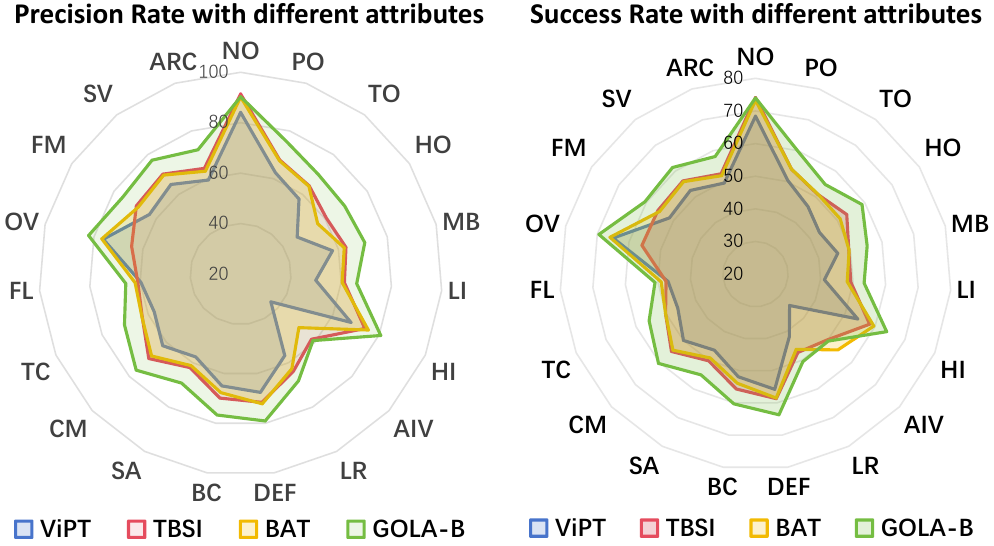}
    \caption{Comparison between GOLA-B with different trackers across various attributes in the LasHeR testing set.}
    \label{fig:attribute}
\end{figure}

\noindent \textbf{Attribute-based Performance.} LasHeR encompasses 19 distinct attributes that serve to evaluate tracker performance across various challenges as depicted in \figref{attribute}. GOLA consistently achieves optimal results across nearly all tested attributes. Notably, within the ``HI" attribute, GOLA's PR exceeds that of TBSI by 6.9\%. Similarly, GOLA also demonstrates outstanding performance on the ``HO" and ``SV" attributes. These results convincingly demonstrate that GOLA is capable of acquiring diverse knowledge for various attributes, thereby enhancing tracking performance.

\begin{table}[!t]
    \centering
    \small
    \setlength{\tabcolsep}{1mm}
    \begin{tabular}{l|cccccc}
        \toprule
            \multirow{2}*{Method} & Params & Trainable & PR & SR & FLOPs & Speed \\
            & (M) & (\%) & (\%) & (\%) & (G) & (\textit{fps}) \\
        \midrule
            FFT & 88 & 100 & 72.5 & 57.9 & 77 & 125 \\
        \midrule
            Adapter & 89 & 4 & 68.8 & 54.5 & 78 & 78 \\
            VPT & 88 & 3 & 70.8 & 56.3 & 78 & 85 \\
            (IA)$^3$ & 88 & 3 & 70.1 & 55.4 & 77 & 80 \\
        \midrule
            LoRA & 99 & 13 & 76.3 & 60.7 & 85 & 125  \\
            AdaLoRA & 99 & 13 & 57.9 & 44.4 & 85 & 125  \\
            DoRA & 99 & 13 & 63.7 & 49.3 & 83 & 125  \\
        \midrule
            GOLA-B & 99 & 10 & 77.5 & 61.6 & 85 & 125 \\
        \bottomrule
    \end{tabular}
    \caption{Comparison between different fine-tuning methods.}
    \label{tab:fft_method}
\end{table}

\noindent \textbf{Comparison between Different Fine-Tuning Methods.} \tabref{fft_method} compares the performance of different fine-tuning methods. The full fine-tuning method achieves 72.5\% of PR and 57.9\% of SR. Fine-tuning methods such as Adapter \cite{adapter}, VPT \cite{vpt}, and (IA)$^3$ \cite{ia3} reduce the proportion of trainable parameters, but their performance is lower compared to full fine-tuning. LoRA \cite{lora} maintains the same inference speed as the full fine-tuning method. DoRA \cite{dora} and AdaLoRA \cite{adalora} require a longer time to fit due to the introduction of new parameters. Compared to LoRA, our method achieves significant performance improvements while reducing the number of trainable parameters by 23\%.

\subsection{Ablation Studies}

\begin{table}[!t]
    \centering
    \small
    \begin{tabular}{c|cc}
        \toprule
            Reference & PR (\%) & SR (\%) \\
        \midrule
            $\mathbf{A}$ & 77.0 & 61.1 \\
            $\mathbf{B}$ & 77.5 & 61.6 \\
        \bottomrule
    \end{tabular}
    \caption{Impact of partitioning reference on performance.}
    \label{tab:reference}
\end{table}

\noindent \textbf{Impact of Partitioning Reference on Performance.} \tabref{reference} demonstrates the impact of different matrices used as sorting criteria on final performance. The experimental results confirm that matrix $\mathbf{B}$ exhibits a stronger correlation with the specific task, thus enabling a more accurate identification and preservation of crucial feature information.

\begin{table}[!t]
    \centering
    \small
    \begin{tabular}{cc|cc}
        \toprule
            Sorting & Clustering & PR (\%) & SR (\%) \\
        \midrule
            $\checkmark$ & - & 77.0 & 61.4 \\
            - & $\checkmark$ & 76.6 & 61.0 \\
        \midrule
            $\checkmark$ & $\checkmark$ & 77.5 & 61.6 \\
        \bottomrule
    \end{tabular}
    \caption{Ablation on rank sorting and clustering strategy.}
    \label{tab:sort_group}
\end{table}

\noindent \textbf{Ablation on Rank Sorting and Grouping Strategy.} In \tabref{sort_group}, we conduct ablations on rank sorting and grouping strategies. The results indicate that rank ordering helps the model retain its generalization ability. At the same time, the grouping strategy facilitates the learning of specific knowledge within groups, thereby enhancing the expressive capability of each group. The combination of the two strategies enhances model performance more effectively.

\begin{table}[!t]
    \centering
    \small
    \begin{tabular}{c|cc}
        \toprule
            Type & PR (\%) & SR (\%) \\
        \midrule
            \ding{172} & 76.3 & 60.7 \\
            \ding{173} & 76.8 & 61.2 \\
            \ding{174} & 76.7 & 61.2 \\
        \midrule
            GOLA-B & 77.5 & 61.6 \\
        \bottomrule
    \end{tabular}
    \caption{Ablation of orthogonal constraint.}
    \label{tab:orth}
\end{table}

\noindent \textbf{Effectiveness of Orthogonal Constraint.} In \tabref{orth}, we explore the effectiveness of our orthogonal constraint applied separately to $\mathbf{A}$ and $\mathbf{B}$. \ding{172} indicates the absence of orthogonal loss, \ding{173} signifies the application of orthogonal loss solely to the matrix $\mathbf{A}$, and \ding{174} denotes the application of orthogonality loss solely to the matrix $\mathbf{B}$. The results indicate that applying the orthogonal loss simultaneously to both $\mathbf{A}$ and $\mathbf{B}$ leads to a more significant enhancement. This suggests the presence of a complementary effect between $\mathbf{A}$ and $\mathbf{B}$, thereby validating the effectiveness of our orthogonal loss.

\begin{table}[!t]
    \centering
    \small
    \begin{tabular}{c|ccc}
        \toprule
            Pair(s) & PR (\%) & SR (\%) & Training Time (m) \\
        \midrule
            1 & 77.5 & 61.6 & 110 \\
            2 & 77.2 & 61.4 & 120 \\
            4 & 77.2 & 61.5 & 150 \\
            8 & 77.0 & 61.3 & 200 \\
        \bottomrule
    \end{tabular}
    \caption{Ablation of sampling quantity on orthogonal loss.}
    \label{tab:pairs}
\end{table}

\noindent \textbf{Impact of Sampling Quantity.} In \tabref{pairs}, we investigate the impact of the number of parameter pairs sampled randomly for the orthogonal constraint. Experimental results indicate that excessive sampling does not yield significant performance improvements. Instead, it increases computational burden. This occurs because a single set of random samples is sufficient to achieve orthogonal constraints. As feature decoupling reaches a certain extent, the effect of adding further constraints shows diminishing returns.

\begin{table}[!t]
    \centering
    \small
    \begin{adjustbox}{width=\linewidth}
        \begin{tabular}{l|ccccc}
            \toprule
                \multirow{2}*{Method} & Params & PR & SR & FLOPs & Speed \\
                & (M) & (\%) & (\%) & (G) & (\textit{fps}) \\
            \midrule
                TBSI & 193 & 61.1 & 46.6 & 97 & 35 \\
                BAT & 94 & 75.1 & 59.3 & 74 & 40 \\
                SeqTrackv2-B256 & 99 & 32.7 & 25.7 & 40 & 46 \\
                SUTrack-B224 & 89 & 71.8 & 56.8 & 70 & 35 \\
            \midrule
                GOLA-B & 99 & 77.5 & 61.6 & 85 & 125 \\
            \bottomrule
        \end{tabular}
    \end{adjustbox}
    \caption{Comparison between different trackers on DINOv2.}
    \label{tab:fusion}
\end{table}

\begin{figure}[!t]
    \centering
    \includegraphics[width=\linewidth]{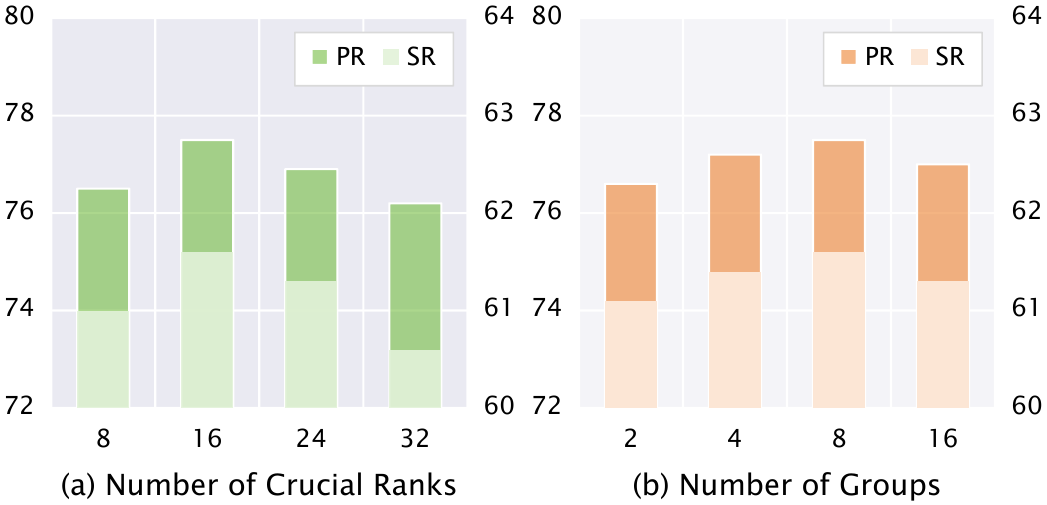}
    \caption{Impact of the number of crucial ranks and groups.}
    \label{fig:group}
\end{figure}

\begin{figure*}[!t]
    \centering
    \includegraphics[width=0.95\linewidth]{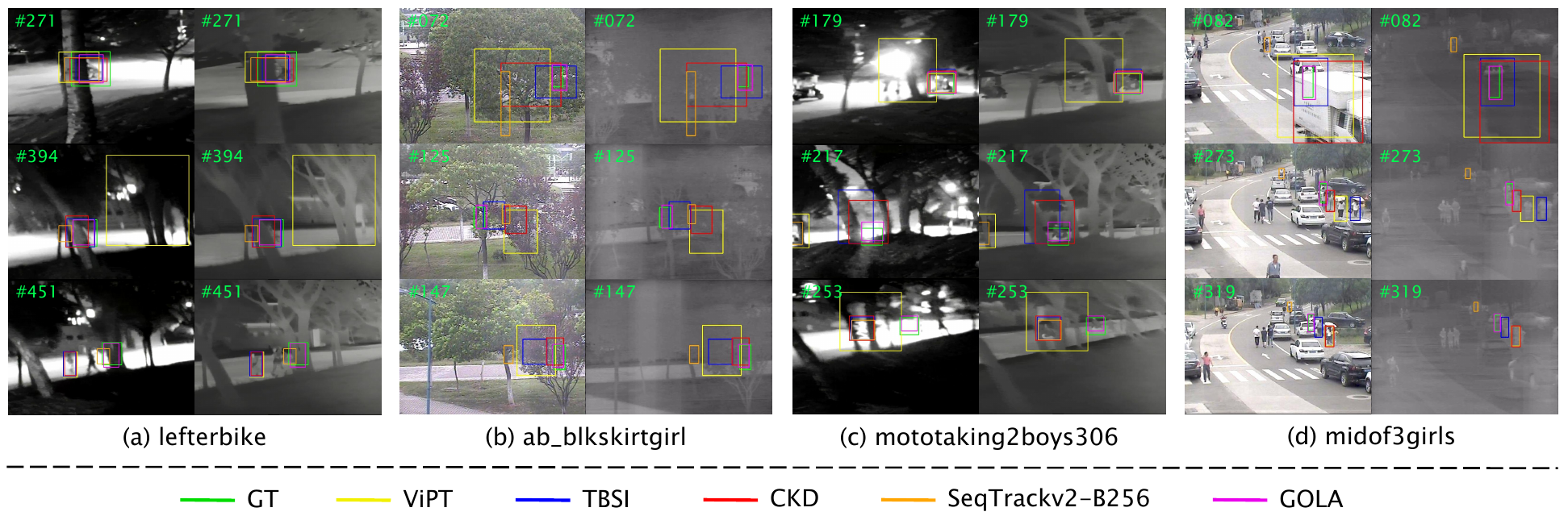}
    \caption{Qualitative comparison of GOLA-B against 4 state-of-the-art trackers on 4 video sequences.}
    \label{fig:visualization}
\end{figure*}

\begin{figure}[!t]
    \centering
    \includegraphics[width=\linewidth]{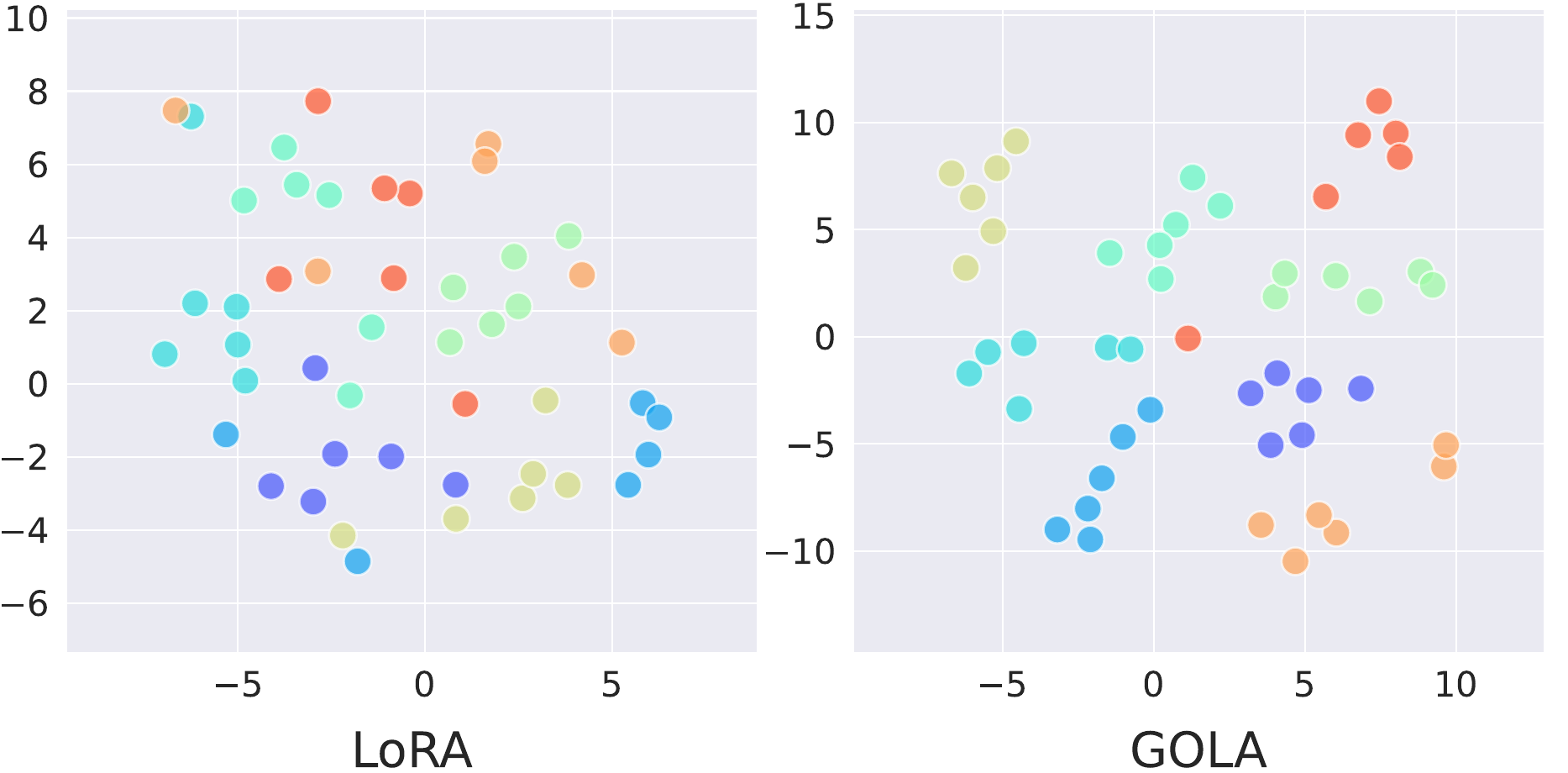}
    \caption{Visualization of t-SNE maps between rank groups. Ranks within different groups use different colors.}
    \label{fig:tsne}
\end{figure}

\noindent \textbf{Performance of Trackers using DINOv2.} To eliminate the impact of different backbones, as shown in \tabref{fusion}, we apply the LoRAT pretrained DINOv2 \cite{dinov2} to other methods to achieve a fairer comparison. All methods adhere to the same training protocol as our model. The new parameters introduced by TBSI and SeqTrackv2 fail to converge effectively within 10 epochs, resulting in suboptimal tracking performance. The full fine-tuning strategy employed by SUTrack leads to overfitting. Although BAT demonstrates superior tracking performance under DINOv2, its dual-stream architecture and additional adapter parameters limit inference speed. In contrast, GOLA maintains optimal tracking performance while exhibiting remarkably high inference speed.

\noindent \textbf{Ablation on Crucial Rank Number.} We conduct an experiment on the number of crucial ranks $k$ as shown in \figref{group} (a). When $k$ is excessively large, abundant redundant ranks become frozen, resulting in insufficient capacity for the model to learn new features. Conversely, an excessively small $k$ fails to retain pre-trained knowledge and impairs the model's original generalization ability.

\noindent \textbf{Ablation on Group Number.} As shown in \figref{group} (b), we present an experiment on the number of groups $n$. The variable $n$ directly influences the expressive capability of the model. As $n$ increases, the number of ranks within each group decreases, resulting in a diminished expressive ability for individual groups. Conversely, as $n$ decreases, having too few groups restricts the model's ability to learn diverse knowledge, ultimately leading to a decline in performance.

\subsection{Visualization and Analysis}

\noindent \textbf{Qualitative Comparison.} We qualitatively compare our method with other RGB-T trackers. As shown in \figref{visualization}, we select four representative sequences from LasHeR, covering diverse tracking challenges such as partial and total occlusions, low illumination, and similar-object interference, to assess performance in realistic scenarios visually. Taking the third sequence as an example, our method demonstrates robust tracking capabilities even in the presence of severe dynamic blur and varying illumination. Results indicate that GOLA learns specific knowledge tailored to different challenges, thereby enhancing tracking performance.

\noindent \textbf{Visualization of Rank Groups.} We visualize the ranks of different groups using t-SNE in \figref{tsne}. For LoRA, we identify the corresponding ranks for visualization by utilizing the grouping indices from GOLA. Compared to LoRA, the rank feature points within the same group of GOLA exhibit a high degree of clustering. This suggests intra-group ranks form a stronger synergistic effect during learning, enabling concentrated expression around specific tasks. Rank orders between different groups show distinct separation, indicating that the groups have learned complementary expressive capabilities. Visualization results further validate the superiority of our approach in learning strategies.

\section{Conclusion}

In this paper, we propose a Group Orthogonal Low-Rank Adaptation (GOLA) framework for RGB-T tracking, aiming to address the limitations in parameter redundancy and low-rank space expression capabilities. GOLA identifies and retains crucial ranks by introducing a rank decomposition partitioning strategy, thereby preserving the priors of the pretrained weights while clustering redundant ranks for structured learning. Furthermore, we implement an inter-group orthogonal constraint strategy to mitigate information redundancy within low-rank spaces, enhancing the model's expressive capabilities for addressing diverse challenges. Compared to full fine-tuning, GOLA reduces the number of trainable parameters to alleviate overfitting and achieves parameter merging during inference without incurring additional delays. Experimental results validate the effectiveness and efficiency of GOLA, demonstrating that our method significantly outperforms state-of-the-art methods while maintaining real-time performance. We hope that GOLA can provide a new perspective for fine-tuning methods in RGB-T tracking and facilitate further advancements in this field.
 
\section{Appendix}

In this appendix, we provide additional material to supplement the details:

\begin{itemize}
    \item Detailed Settings
    \item More Ablation Studies
    \item Further Analysis
    \item Limitation
\end{itemize}

\section{Detailed Settings}

\subsection{Computing Environments}
Our experiments are conducted on an Ubuntu 22.04 server equipped with an Intel(R) Xeon(R) Platinum 8276 CPU @ 2.20GHz and 768GB of RAM. This server is also outfitted with four NVIDIA A40 GPUs, each possessing 48GB of VRAM for high-speed acceleration. In terms of software configuration, we utilize Python 3.10.15, PyTorch 2.5.1, and CUDA 11.8 to support computational operations.

\subsection{Variant Details}
We demonstrate the detailed parameters of different variants in \tabref{variant_details}. Considering the varying demands for model performance and efficiency across different application scenarios, we design two variants, GOLA-B and GOLA-L, to validate the performance of our method under different backbone scales. Specifically, GOLA-B adopts DINOv2-B224 \cite{dinov2} as the backbone to balance performance and computational efficiency, making it suitable for resource-constrained real-time tracking scenarios. GOLA-L selects DINOv2-L224 as the backbone to enhance feature extraction capabilities for addressing more complex multi-modal interference scenarios. The two variants differ only in backbone scale, while maintaining consistent rank decomposition partition strategies and inter-group orthogonal constraint mechanisms to verify the generalizability and stability of our method across different backbone configurations.

\subsection{Datasets}


\noindent \textbf{GTOT \cite{gtot}.} The GTOT dataset represents the initial benchmark meticulously crafted for RGB-T tracking. It encompasses 50 precisely paired video sequences, each containing an infrared video and its associated grayscale video. The videos in this dataset were captured across diverse settings, including laboratories, campus thoroughfares, and playgrounds. As such, GTOT provides a comprehensive set of scenarios for evaluating various trackers.


\noindent \textbf{RGBT210 \cite{rgbt210}.} The RGBT210 dataset contains 210 RGB and infrared video pairs, all precisely annotated. With over 210,000 frames in total, it provides ground-truth annotations for every tracked object. The dataset is instrumental because it identifies 12 distinct attributes like object deformation and fast motion. These attributes enable a nuanced performance evaluation of trackers by considering specific object behaviours and motions.


\noindent \textbf{RGBT234 \cite{rgbt234}.} The RGBT234 dataset consists of 234 precisely aligned pairs of visible and infrared video sequences spanning around 234,000 frames. Some individual sequences can be as long as approximately 8,000 frames. This dataset significantly broadens the data availability. It also places a strong emphasis on offering detailed annotation. Each entry has up to 12 distinct attributes, enabling more in-depth data analysis.


\noindent \textbf{LasHeR \cite{lasher}.} The LasHeR dataset is a large-scale RGB-T tracking dataset, encompassing 1,224 pairs of visible and infrared videos, totalling over 734,800 frame pairs. All frame pairs are spatially aligned and manually annotated with bounding boxes, ensuring high-quality and densely annotated data. The dataset encompasses a wide range of object categories, multiple camera perspectives, and diverse environmental factors, including various seasons, weather conditions, and day-night variations, thereby showcasing a high degree of diversity.

\begin{table}[!t]
    \centering
    \small
    \begin{tabular}{c|cc}
        \toprule
            Parameter & GOLA-B & GOLA-L \\
        \midrule
            Patch Size & 14 & 14 \\
            Embed Dim & 768 & 1024 \\
            Head & 12 & 16 \\
            Depth & 12 & 20 \\
        \bottomrule
    \end{tabular}
    \caption{Detailed settings of different variants of our method.}
    \label{tab:variant_details}
\end{table}

\subsection{Evaluation Metrics}

\noindent \textbf{Precision Rate.} The Precision Rate (PR) evaluates the accuracy of predicted object locations. It does so by computing the Euclidean distance between the centres of the tracking bounding box and the ground truth bounding box. Specifically, PR represents the fraction of frames where the distance between the tracking outcome and the object's position is beneath a predefined threshold. Mathematically, PR is calculated using the formula: 
\begin{equation}
    \text{PR}=\frac{1}{N}\sum^N_{t=1}\delta(c_t < \xi_{pr}).
\end{equation}
In this formula, $c_t$ represents the distance between the predicted and actual center locations in frame $t$. $N$ refers to the overall number of frames. $\delta(\cdot)$ functions as the indicator function, while $\xi_{pr}$ is the threshold value.

\noindent \textbf{Success Rate.} The Success Rate (SR) gauges the overlap degree between the predicted bounding boxes and their ground-truth counterparts. It is calculated as the proportion of frames where the Intersection over Union (IoU) surpasses a given threshold. Mathematically, SR is formulated as:
\begin{equation}
    \text{SR}=\frac{1}{N}\sum^N_{t=1}\delta(\text{IoU}_t\ge \xi_{sr}).
\end{equation}
Here, $\text{IoU}_t$ denotes the IoU metric computed between the predicted and actual bounding boxes within frame $t$. $\xi_{sr}$ represents the predefined threshold value used for assessment.

\noindent \textbf{Maximum Precision Rate.} The Maximum Precision Rate (MPR) is designed to identify alignment errors in RGB and infrared images. It assesses precision by calculating the minimum distance between the centres of the ground truth and predicted boxes across both visible and infrared datasets.

\noindent \textbf{Maximum Success Rate.} The Maximum Success Rate (MSR) is crafted to tackle alignment errors in RGB and infrared images. Instead of relying on just one dataset, MSR gauges the success rate by considering the larger IoU calculated between the ground truth and predicted boxes from visible and infrared imagery. This approach helps in getting a more comprehensive measure of alignment performance.

\section{More Ablation Studies}

\begin{table}[!t]
    \centering
    \small
    \begin{tabular}{c|cc}
        \toprule
            $\lambda$ & PR (\%) & SR (\%) \\
        \midrule
            $1.0 \times 10^{-3}$ & 77.0 & 61.2 \\
            $1.2 \times 10^{-3}$ & 77.2 & 61.4 \\
            $1.4 \times 10^{-3}$ & \textbf{77.5} & \textbf{61.6} \\
            $1.6 \times 10^{-3}$  & 76.6 & 61.0 \\
        \bottomrule
    \end{tabular}
    \caption{Ablation on the balance factor $\lambda$.}
    \label{tab:lambda}
\end{table}

\noindent \textbf{Ablation on the Balance Factor $\lambda$.} In \tabref{lambda}, we conduct an experiment on the balance factor $\lambda$ of the orthogonal loss. Results show when $\lambda = 1.4 \times 10^{-3}$, the model achieves optimal performance, demonstrating that the orthogonal constraint effectively enhances the model's generalization capability. Further increasing or decreasing $\lambda$ compromises the representation space, leading to performance decline.

\begin{table}[!t]
    \centering
    \small
    \begin{tabular}{c|cc}
        \toprule
            $\tau$ & PR (\%) & SR (\%) \\
        \midrule
            0.83 & 77.3 & 61.5 \\
            0.84 & \textbf{77.5} & \textbf{61.6} \\
            0.85 & 77.0 & 61.3 \\
            0.86  & 76.1 & 60.6 \\
        \bottomrule
    \end{tabular}
    \caption{Ablation on the template update threshold $\tau$.}
    \label{tab:tau}
\end{table}

\noindent \textbf{Ablation on the Template Update Threshold $\tau$.} We investigate the impact of the template update threshold $\tau$ on the performance of our proposed method. As shown in \tabref{tau}, we vary $\tau$ between 0.83 and 0.86. When $\tau$ is too low, the method tends to update templates more frequently, which can potentially introduce noise. Conversely, when $\tau$ is higher, the template updates become more conservative, which may limit the adaptability of the tracker.

\begin{table}[!t]
    \centering
    \small
    \begin{tabular}{c|cc}
        \toprule
            Online Template & PR (\%) & SR (\%) \\
        \midrule
            - & 73.5 & 58.6 \\
            \checkmark & \textbf{77.5} & \textbf{61.6} \\
        \bottomrule
    \end{tabular}
    \caption{Ablation of online template on GOLA-B.}
    \label{tab:ot}
\end{table}

\noindent \textbf{Ablation of Online Template.} We conduct an ablation of the online template in \tabref{ot}. The results show that introducing the online template brings a clear performance gain. As a widely adopted strategy, the online template plays a crucial role in alleviating variations in appearance and pose. Notably, even without using the online template, our method still outperforms other approaches that also do not employ the online template, such as TBSI, BAT, and CKD.

\section{Further Analysis}

\begin{table}[!t]
    \centering
    \small
    \begin{tabular}{c|cccc|cc}
        \toprule
            Attribute & FFT & LoRA & GOLA-B \\ 
        \midrule
            NO & 90.7/73.3 & \textbf{91.2}/\textbf{74.9} & 90.3/74.0 \\ 
            PO & 70.0/56.0 & 74.3/58.8 & \textbf{75.7}/\textbf{59.9} \\ 
            TO & 65.6/51.6 & 68.7/54.1 & \textbf{70.1}/\textbf{54.8} \\ 
            HO & 57.8/51.8 & 69.0/58.9 & \textbf{69.3}/\textbf{59.0} \\ 
            MB & 65.5/51.6 & 69.9/54.9 & \textbf{70.8}/\textbf{55.3} \\ 
            LI & 60.8/49.6 & \textbf{66.7}/\textbf{53.8} & 66.2/53.4 \\ 
            HI & 80.0/62.9 & 79.8/63.7 & \textbf{80.7}/\textbf{63.9} \\ 
            AIV & 56.9/49.4 & 57.4/49.1 & \textbf{59.0}/\textbf{50.3} \\ 
            LR & 65.2/47.7 & 67.3/50.0 & \textbf{68.2}/\textbf{50.5} \\ 
            DEF & 73.3/59.2 & 75.1/60.6 & \textbf{79.1}/\textbf{63.7} \\ 
            BC & 71.9/57.0 & 74.9/59.2 & \textbf{76.7}/\textbf{60.3} \\ 
            SA & 63.4/50.7 & 66.7/53.3 & \textbf{69.2}/\textbf{55.1} \\ 
            CM & 71.2/56.5 & 75.1/59.6 & \textbf{76.3}/\textbf{60.4} \\ 
            TC & 65.3/51.8 & 68.8/54.6 & \textbf{70.3}/\textbf{55.6} \\ 
            FL & 63.3/49.9 & \textbf{66.0}/\textbf{51.5} & 65.7/50.9 \\ 
            OV & 70.6/62.2 & 72.3/62.2 & \textbf{82.2}/\textbf{69.5} \\ 
            FM & 70.8/57.1 & 75.2/60.1 & \textbf{75.8}/\textbf{60.5} \\ 
            SV & 72.4/58.0 & 76.0/60.5 & \textbf{77.2}/\textbf{61.4} \\ 
            ARC & 67.2/54.4 & 71.2/57.2 & \textbf{72.1}/\textbf{58.1} \\ 
        \midrule
            ALL & 72.5/57.9 & 76.3/60.7 & \textbf{77.5}/\textbf{61.6} \\ 
        \bottomrule
    \end{tabular}
    \caption{Attribute-based PR/SR scores (\%) on the LasHeR dataset compared with several fine-tuning methods. The best results are highlighted in \textbf{bold}.}
    \label{tab:attribute_lasher_lora}
\end{table}

\begin{table*}[!t]
    \centering
    \small
    \begin{tabular}{c|ccccc|c}
        \toprule
            Attribute & APFNet & ViPT & TBSI & CKD & STTrack & GOLA-B \\ 
        \midrule
            NO & 66.7/46.7 & 84.1/68.4 & 91.4/74.1 & \textbf{91.6}/73.9 & 91.3/\textbf{74.4} & 90.3/74.0 \\ 
            PO & 47.3/34.5 & 62.4/50.3 & 67.8/54.0 & 70.7/56.0 & 74.1/58.7 & \textbf{75.7}/\textbf{59.9} \\ 
            TO & 41.7/31.4 & 57.6/46.1 & 64.3/51.0 & 66.9/52.7 & 67.4/52.8 & \textbf{70.1}/\textbf{54.8} \\ 
            HO & 27.1/27.7 & 43.7/43.8 & 60.6/53.4 & 64.2/55.6 & 57.5/51.2 & \textbf{69.3}/\textbf{59.0} \\ 
            MB & 45.9/32.8 & 57.3/45.9 & 63.1/49.5 & 67.6/52.9 & 69.0/54.3 & \textbf{70.8}/\textbf{55.3} \\ 
            LI & 41.8/30.8 & 49.8/41.2 & 61.3/49.3 & 64.1/51.0 & 65.3/51.5 & \textbf{66.2}/\textbf{53.4} \\ 
            HI & 60.4/41.2 & 67.9/54.2 & 73.8/58.2 & 78.9/61.5 & 78.9/62.8 & \textbf{80.7}/\textbf{63.9} \\ 
            AIV & 32.1/26.2 & 37.5/35.0 & 58.2/49.8 & \textbf{61.2}/\textbf{52.0} & 55.6/49.2 & 59.0/50.3 \\ 
            LR & 46.1/29.4 & 56.4/41.6 & 63.9/47.3 & 64.3/46.9 & \textbf{70.5}/\textbf{51.9} & 68.2/50.5 \\ 
            DEF & 45.8/36.8 & 67.4/55.7 & 71.6/58.7 & 76.8/62.0 & 76.6/62.0 & \textbf{79.1}/\textbf{63.7} \\ 
            BC & 44.9/33.7 & 64.9/51.8 & 69.9/55.7 & 71.3/56.3 & 74.7/59.0 & \textbf{76.7}/\textbf{60.3} \\ 
            SA & 42.8/31.7 & 57.3/46.5 & 62.2/50.2 & 64.7/51.8 & \textbf{69.9}/\textbf{55.4} & 69.2/55.1 \\ 
            CM & 47.7/35.1 & 62.1/50.0 & 69.5/55.0 & 72.1/56.9 & 73.1/58.2 & \textbf{76.3}/\textbf{60.4} \\ 
            TC & 43.1/31.6 & 57.3/46.0 & 62.6/50.1 & 66.3/52.4 & 68.6/54.3 & \textbf{70.3}/\textbf{55.6} \\ 
            FL & 37.6/27.9 & 59.1/46.5 & 60.9/47.5 & 66.3/51.8 & \textbf{66.8}/\textbf{53.1} & 65.7/50.9 \\ 
            OV & 36.4/34.2 & 76.2/65.0 & 64.6/55.9 & 72.7/62.4 & 72.1/62.2 & \textbf{82.2}/\textbf{69.5} \\ 
            FM & 45.1/33.9 & 63.1/51.4 & 69.4/55.7 & 72.0/57.4 & 74.2/59.3 & \textbf{75.8}/\textbf{60.5} \\ 
            SV & 49.8/36.0 & 65.0/52.5 & 70.2/56.2 & 73.0/58.0 & 76.4/60.7 & \textbf{77.2}/\textbf{61.4} \\ 
            ARC & 40.5/31.0 & 59.3/49.5 & 64.3/52.5 & 68.0/55.0 & 70.9/57.6 & \textbf{72.1}/\textbf{58.1} \\ 
        \midrule
            ALL & 50.0/36.2 & 65.1/52.5 & 70.5/56.3 & 73.2/58.1 & 76.0/60.4 & \textbf{77.5}/\textbf{61.6} \\ 
        \bottomrule
    \end{tabular}
    \caption{Attribute-based PR/SR scores (\%) on the LasHeR dataset compared with several state-of-the-art methods. The best results are highlighted in \textbf{bold}.}
    \label{tab:attribute_lasher}
\end{table*}

\begin{table*}[!t]
    \centering
    \small
    \begin{tabular}{c|ccc|ccc}
        \toprule
            Attribute & SeqTrackv2-B256 & SeqTrackv2-B384 & GOLA-B & SeqTrackv2-L256 & SeqTrackv2-L384 & GOLA-L \\ 
        \midrule
            NO & 89.2/72.0 & 86.2/69.8 & \textbf{90.3}/\textbf{74.0} & 89.8/74.4 & \textbf{91.6}/\textbf{75.3} & 90.2/74.1 \\ 
            PO & 67.6/53.5 & 69.2/54.2 & \textbf{75.7}/\textbf{59.9} & 71.9/56.7 & 74.7/59.1 & \textbf{76.4}/\textbf{60.3} \\ 
            TO & 64.0/50.1 & 63.1/49.3 & \textbf{70.1}/\textbf{54.8} & 65.3/51.4 & 68.6/54.0 & \textbf{71.4}/\textbf{55.8} \\ 
            HO & 53.9/47.4 & 54.8/48.2 & \textbf{69.3}/\textbf{59.0} & 63.0/54.5 & 64.2/55.9 & \textbf{68.3}/\textbf{58.1} \\ 
            MB & 65.6/51.6 & 65.0/50.4 & \textbf{70.8}/\textbf{55.3} & 67.8/53.1 & 71.7/56.2 & \textbf{73.5}/\textbf{57.5} \\ 
            LI & 59.9/47.4 & 59.4/47.0 & \textbf{66.2}/\textbf{53.4} & 62.9/49.9 & 68.0/53.8 & \textbf{69.1}/\textbf{55.0} \\ 
            HI & 77.2/60.5 & 80.2/62.6 & \textbf{80.7}/\textbf{63.9} & 79.6/63.1 & \textbf{84.0}/\textbf{66.6} & \textbf{84.0}/66.1 \\ 
            AIV & 48.5/43.5 & 41.5/36.9 & \textbf{59.0}/\textbf{50.3} & 54.1/46.6 & 61.9/52.7 & \textbf{62.6}/\textbf{53.7} \\ 
            LR & 62.9/46.3 & 66.1/47.7 & \textbf{68.2}/\textbf{50.5} & 66.5/48.8 & 69.3/51.3 & \textbf{69.8}/\textbf{52.3} \\ 
            DEF & 74.5/59.6 & 70.7/57.2 & \textbf{79.1}/\textbf{63.7} & 73.4/59.1 & 77.0/62.5 & \textbf{78.5}/\textbf{62.7} \\ 
            BC & 69.0/54.1 & 70.1/54.6 & \textbf{76.7}/\textbf{60.3} & 70.1/54.8 & 74.5/58.8 & \textbf{78.1}/\textbf{61.8} \\ 
            SA & 59.8/47.8 & 62.3/49.1 & \textbf{69.2}/\textbf{55.1} & 64.9/51.6 & 67.3/54.0 & \textbf{69.1}/\textbf{54.6} \\ 
            CM & 68.7/53.9 & 69.6/54.3 & \textbf{76.3}/\textbf{60.4 }& 72.4/57.4 & 75.8/60.2 & \textbf{79.0}/\textbf{62.4} \\ 
            TC & 61.5/48.6 & 63.7/49.8 & \textbf{70.3}/\textbf{55.6} & 65.3/51.7 & 69.2/55.0 & \textbf{71.6}/\textbf{56.6} \\ 
            FL & 65.1/50.6 & 63.1/49.5 & \textbf{65.7}/\textbf{50.9} & 71.6/55.4 & \textbf{75.5}/\textbf{59.2} & 68.6/53.2 \\ 
            OV & 78.1/66.7 & 78.8/66.7 & \textbf{82.2}/\textbf{69.5} & 77.1/66.3 & \textbf{84.4}/\textbf{72.7} & 68.4/58.7 \\ 
            FM & 68.9/55.0 & 69.9/55.3 & \textbf{75.8}/\textbf{60.5} & 73.5/58.6 & 76.1/60.9 & \textbf{76.8}/\textbf{61.3} \\ 
            SV & 70.8/56.3 & 71.3/56.2 & \textbf{77.2}/\textbf{61.4} & 74.0/58.9 & 76.3/60.8 & \textbf{78.2}/\textbf{62.0} \\ 
            ARC & 64.5/52.5 & 63.8/51.6 & \textbf{72.1}/\textbf{58.1} & 68.6/55.5 & 72.0/58.0 & \textbf{72.7}/\textbf{58.2} \\ 
        \midrule
            ALL & 70.4/55.8 & 71.5/56.2 & \textbf{76.8}/\textbf{61.3} & 74.1/58.8 & 76.7/61.0 & \textbf{78.1}/\textbf{61.9} \\ 
        \bottomrule
    \end{tabular}
    \caption{Attribute-based PR/SR scores (\%) on LasHeR compared with SeqTrackv2. The best results are highlighted in \textbf{bold}.}
    \label{tab:attribute_seqtrackv2}
\end{table*}

\begin{table*}[!t]
    \centering
    \small
    \begin{tabular}{c|ccccc|c} 
        \toprule
        Attribute & ViPT & TBSI & MPLT & CKD & STTrack & GOLA-B \\
        \midrule
        NO & 92.4/71.0 & 96.2/73.9 & 97.7/74.8 & \textbf{97.9}/75.7 & 97.6/74.6 & 97.6/\textbf{75.9} \\
        PO & 85.4/63.0 & 88.6/66.1 & 88.5/65.5 & 90.6/68.1 & 89.4/66.4 & \textbf{93.5}/\textbf{70.4} \\
        HO & 77.6/56.3 & 83.8/61.9 & 84.1/61.7 & 86.0/63.1 & 86.7/63.4 & \textbf{88.6}/\textbf{65.7} \\
        LI & 81.0/58.4 & 88.3/65.2 & 87.6/64.4 & 91.9/67.4 & 89.9/65.5 & \textbf{94.8}/\textbf{70.5} \\
        LR & 83.1/59.4 & 85.6/62.2 & 87.6/62.8 & 86.7/62.0 & 84.2/60.7 & \textbf{89.7}/\textbf{65.4} \\
        TC & 83.0/62.2 & 86.2/65.1 & 85.0/64.2 & 87.4/66.0 & \textbf{91.5}/68.0 & 89.9/\textbf{66.6} \\
        DEF & 81.7/62.2 & 84.6/65.0 & 85.8/65.4 & 86.6/66.5 & 86.5/65.3 & \textbf{90.1}/\textbf{69.2} \\
        FM & 80.3/58.6 & 82.3/61.1 & 83.5/61.8 & 88.0/64.7 & 84.6/62.5 & \textbf{88.6}/\textbf{64.5} \\
        SV & 83.8/63.0 & 89.3/67.6 & 89.0/67.2 & 90.6/68.7 & 92.7/69.3 & \textbf{93.8}/\textbf{71.5} \\
        MB & 83.2/62.6 & 89.2/67.7 & 86.1/64.8 & 91.5/69.8 & 88.6/67.4 & \textbf{90.6}/\textbf{69.7} \\
        CM & 83.0/62.1 & 87.0/66.0 & 86.9/65.3 & 90.7/68.6 & 88.7/66.8 & \textbf{90.7}/\textbf{69.5} \\
        BC & 79.6/55.7 & 83.9/59.9 & 84.5/60.4 & 87.2/62.1 & 85.3/60.4 & \textbf{87.9}/\textbf{63.4} \\
        \midrule
        ALL & 83.5/61.7 & 88.0/65.8 & 88.4/65.7 & 90.0/67.4 & 89.8/66.7 & \textbf{92.2}/\textbf{69.5} \\
        \bottomrule
    \end{tabular}
    \caption{Attribute-based MPR/MSR scores (\%) on RGBT234 compared with several state-of-the-art trackers. The best results are highlighted in \textbf{bold}.}
    \label{tab:attribute_rgbt234}
\end{table*}

\begin{table*}[!t]
    \centering
    \small
    \begin{tabular}{c|cccc|c} 
        \toprule
        Attribute & ViPT & TBSI & MPLT & CKD & GOLA-B \\
        \midrule
        NO & 94.9/73.8 & 94.6/73.8 & 95.4/74.2 & \textbf{96.7}/\textbf{74.0} & 96.4/\textbf{74.0} \\
        PO & 91.1/68.4 & 91.4/68.5 & 90.8/67.7 & 89.3/66.3 & \textbf{93.3}/\textbf{69.2} \\
        HO & \textbf{86.5}/\textbf{62.2} & 85.6/61.2 & 82.9/59.5 & 84.2/60.5 & 86.1/61.9 \\
        LI & 91.7/67.2 & 89.2/65.5 & 88.8/65.4 & 89.3/64.2 & \textbf{94.9}/\textbf{69.2} \\
        LR & 76.3/52.7 & 81.9/55.9 & 84.6/57.7 & 79.6/54.0 & \textbf{84.7}/\textbf{58.0} \\
        TC & 85.6/64.3 & 87.4/65.3 & 84.5/63.3 & 81.8/61.5 & \textbf{90.4}/\textbf{65.3} \\
        DEF & 88.3/\textbf{66.9} & \textbf{88.8}/66.8 & 88.3/66.6 & 86.2/65.0 & 88.2/66.2 \\
        FM & 83.7/61.2 & 87.1/\textbf{63.3} & \textbf{87.6}/63.2 & 87.2/62.9 & 85.6/60.8 \\
        SV & 90.6/69.1 & 90.5/69.1 & 89.4/68.0 & 90.0/67.7 & \textbf{93.1}/\textbf{69.8} \\
        MB & 84.7/62.9 & 85.2/62.9 & 85.6/62.8 & 87.9/65.1 & \textbf{90.8}/\textbf{67.7} \\
        CM & 86.4/63.9 & 87.8/64.4 & 85.6/62.8 & 87.8/64.8 & \textbf{88.9}/\textbf{66.4} \\
        BC & 82.9/57.9 & 82.3/57.0 & 79.2/54.9 & 85.0/59.2 & \textbf{86.3}/\textbf{60.3} \\
        \midrule
        ALL & 89.9/66.8 & 89.6/66.4 & 88.4/65.4 & 88.4/65.2 & \textbf{90.9}/\textbf{67.0} \\
        \bottomrule
    \end{tabular}
    \caption{Attribute-based PR/SR scores (\%) on the RGBT210 dataset compared with several state-of-the-art trackers. The best results are highlighted in \textbf{bold}.}
    \label{tab:attribute_rgbt210}
\end{table*}

\noindent \textbf{Attribute-based Performance of Fine-Tuning Methods.} Results presented in \tabref{attribute_lasher_lora} highlight the comparative performance of different fine-tuning methods, focusing on the attribute-based scores in the LasHeR dataset. Notably, although both LoRA and GOLA surpass the full fine-tuning (FFT) method in many attributes, GOLA consistently performs better. In particular, GOLA significantly improves attributes such as OV, SA, and BC. GOLA outperforms LoRA in most cases, confirming that our method is better suited for learning specific transformations to enhance tracking performance in response to diverse challenges.

\noindent \textbf{More Attribute-based Comparisons.} We select the LasHeR, RGBT234, and RGBT210 datasets to conduct a detailed comparison across different attributes. In \tabref{attribute_lasher}, we compare our method with five state-of-the-art trackers, APFNet \cite{apfnet}, ViPT \cite{vipt}, TBSI \cite{tbsi}, CKD \cite{ckd}, and STTrack \cite{sttrack}. Results indicate that our approach outperforms STTrack on most attributes. To ensure a fairer comparison across model size dimensions, we perform comparisons in \tabref{attribute_seqtrackv2}, where we evaluate all versions of our method against SeqTrackv2 \cite{seqtrackv2} on the LasHeR test set. Even when compared with the larger SeqTrackv2, our best variant consistently demonstrates superior performance across most attributes, validating the effectiveness and robustness of our approach. Additionally, in \tabref{attribute_rgbt234} and \tabref{attribute_rgbt210}, we conduct comparisons of our method with ViPT, TBSI, MPLT \cite{mplt}, CKD, and STTrack for RGBT234 and RGBT210, respectively. It is noteworthy that our method demonstrates excellent performance across most attributes. This clearly evidences the strong adaptability of the framework we propose, allowing it to effectively tackle the diverse and complex challenges in different scenarios.

\begin{table}[!t]
    \centering
    \small
    \begin{tabular}{c|c|c}
        \toprule
           Methods & Backbone & VRAM (GB) \\
        \midrule
            TBSI & ViT-B256 & 22.9 \\
            MPLT & ViT-B256 & 24.2 \\
            BAT & ViT-B256 & 14.3 \\
        \midrule
            GOLA-B & DINOv2-B224 & 10.5 \\
            GOLA-L & DINOv2-L224 & 25.7 \\
        \bottomrule
    \end{tabular}
    \caption{VRAM usage of trackers under batch size 32.}
    \label{tab:vram}
\end{table}


\noindent \textbf{Memory Efficiency.} As detailed in \tabref{vram}, the GOLA variants maintain a competitive memory footprint given their model capacities. Specifically, GOLA-B requires only 10.5GB of VRAM with a batch size of 32, making it suitable for training on standard GPUs. Although GOLA-L consumes more VRAM, it offers enhanced representational capability and superior tracking accuracy.

\begin{figure}[!t]
    \centering
    \includegraphics[width=\linewidth]{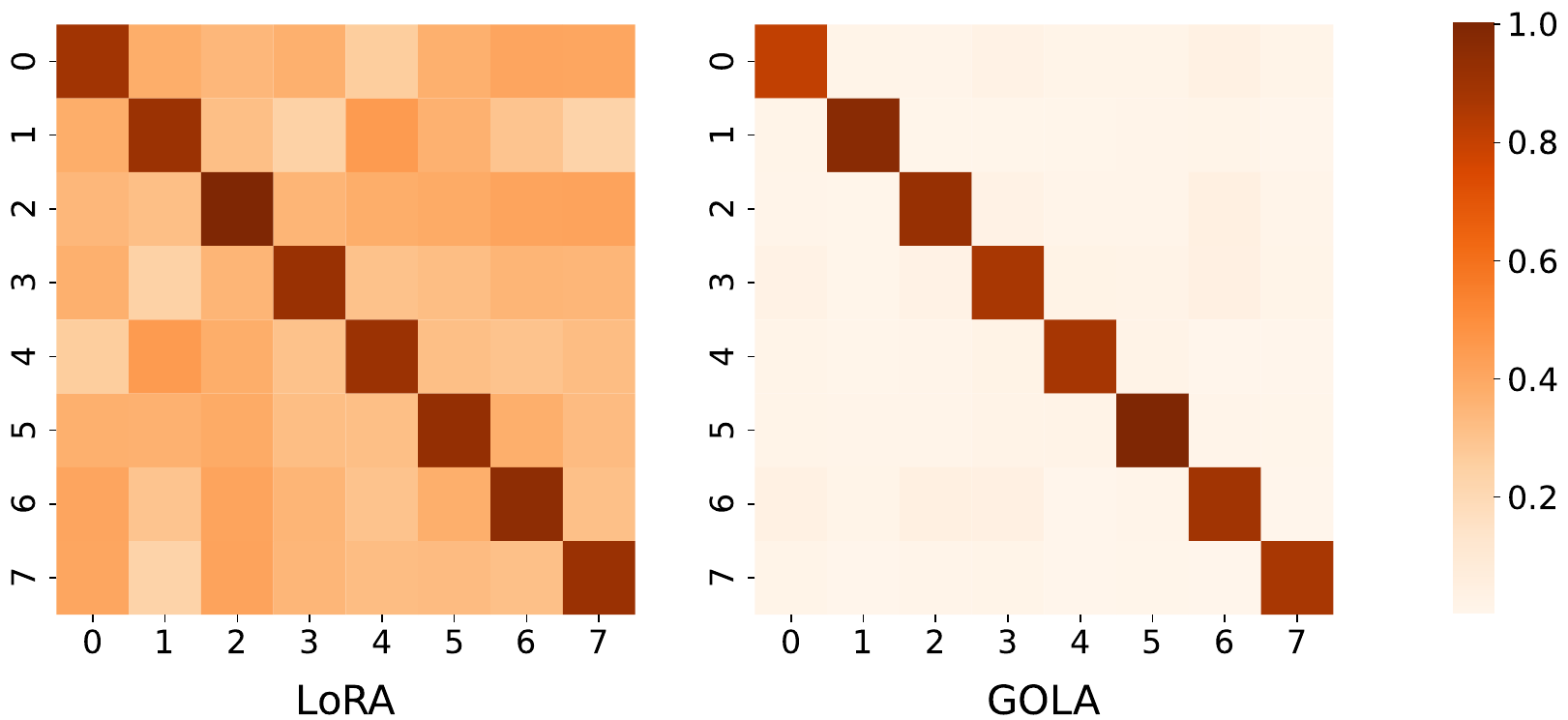}
    \caption{Normalized orthogonal heatmap between groups.}
    \label{fig:orth}
\end{figure}

\noindent \textbf{Visualization of orthogonality between groups.} In \figref{orth}, we show the normalized orthogonality of the low-rank matrices of the model across different groups. For LoRA \cite{lora}, we utilize the group indices from GOLA to identify the corresponding ranks for visualization. This result confirms that the orthogonality between enhanced groups effectively reduces redundancy, allowing different rank groups to focus on learning complementary feature dimensions. This structured partitioning of the feature space aligns with the fundamental principles of efficient representation learning. When the model is compelled to explore diverse knowledge under orthogonal constraints, it can more fully leverage the expressive potential of low-rank spaces, ultimately translating into performance enhancement.

\begin{figure*}[!t]
    \centering
    \includegraphics[width=\linewidth]{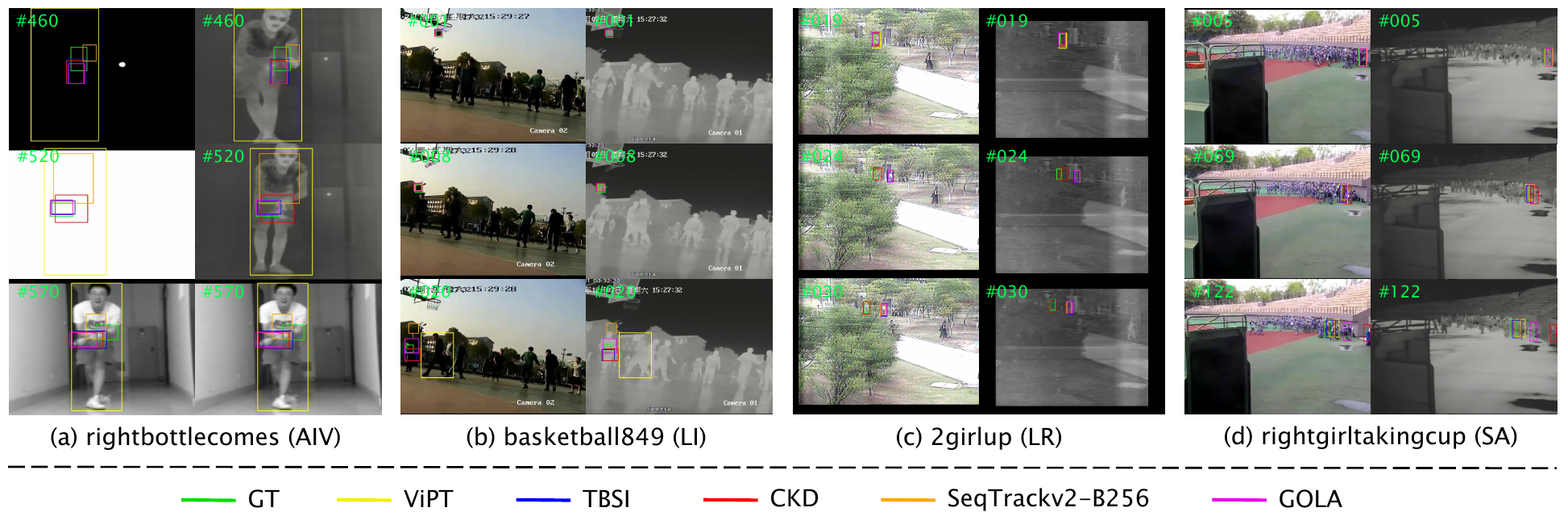}
    \caption{Visualization of failure cases of GOLA-B under 4 representative attributes.}
    \label{fig:failure}
\end{figure*}

\noindent \textbf{Failure Case Analysis.} To better understand the limitations of GOLA, \figref{failure} visualizes failure cases for four attributes with low metric scores, which correspond to abrupt illumination variations (AIV), low illumination (LI), low resolution (LR), and similar appearance (SA). In the AIV case, RGB is affected by sudden strong illumination changes, while thermal data becomes unstable due to sensor noise. GOLA tracks the object briefly but eventually drifts to a nearby region, indicating that frozen crucial ranks lack sufficient adaptability to drastic cross-modal appearance shifts. Under LI, both RGB and thermal images show weak object-background contrast. Though orthogonal rank groups are designed to leverage cross-modal complementary cues, severely degraded modalities cause learned groups to focus on salient yet irrelevant structures, leading to irreversible drift. For LR, the tiny object is surrounded by cluttered distractors. GOLA gradually locks onto a larger neighboring object, revealing that fixed rank budgets and group partitioning lack capacity to model the fine-grained local patterns required for tiny objects. In the SA case, the scene contains multiple objects with highly similar appearance and motion. GOLA maintains the correct object for several frames but eventually switches to a similar distractor. This indicates that despite orthogonal constraint-induced diversity, the current design does not explicitly enforce strong identity discrimination across rank groups, leaving GOLA vulnerable to confusing similar-appearance objects. Overall, these representative failure cases illustrate the multifaceted challenges faced by GOLA in real-world complex scenarios, underscoring the need for more flexible rank adaptation and discriminative feature learning strategies in future iterations.

\noindent \textbf{Discussion on Training without Priors.} GOLA relies on strong pretrained LoRA weights to provide meaningful priors that are then decomposed into orthogonal rank groups. These priors supply rich and diverse feature subspaces, enabling the orthogonal constraint to refine and reorganize complementary knowledge across modalities rather than create it from scratch. When the model is trained without such priors, the orthogonal term is forced to operate on poorly structured, low-quality subspaces. In this case, enforcing orthogonality cannot compensate for the inherent insufficiency of the representations, it merely spreads random or weak features across different groups, effectively degenerating GOLA into a generalized LoRA with an additional orthogonal regularizer. As a result, the learned groups no longer reflect complementary modality-specific or modality-shared patterns, and the model’s behavior deviates from the intended multimodal transfer design.

\section{Limitation}

GOLA exhibits certain limitations in terms of application scope, training efficiency, and hyperparameter sensitivity. In its application scope, GOLA depends on the pretrained weights of LoRA, which restricts its use to LoRA-based fine-tuning methods. Regarding training efficiency, although the orthogonal regularization adopts a random sampling strategy to reduce the computational cost of the orthogonal loss, dense orthogonal computations are still required for different parameter groups, which increases the overall training time to some extent. In addition, GOLA involves a relatively large number of manually specified hyperparameters, and its performance is sensitive to these hyperparameters, making the training process more challenging. Future research aims to extend the proposed method to a broader range of fine-tuning frameworks and to explore strategies such as automatic weight selection and soft diversity regularization to enhance robustness and expand applicability.

\section{Acknowledgments}
This work was supported by the Joint Funds of the National Natural Science Foundation of China (U2441251, U24A20218), in part by the National Natural Science Foundation of China (62303046).

\bibliography{aaai2026}

@inproceedings{lorat,
  title={Tracking meets lora: Faster training, larger model, stronger performance},
  author={Lin, Liting and Fan, Heng and Zhang, Zhipeng and Wang, Yaowei and Xu, Yong and Ling, Haibin},
  booktitle={ECCV},
  pages={300--318},
  year={2024},
  organization={Springer}
}

@inproceedings{tbsi,
  title={Bridging search region interaction with template for rgb-t tracking},
  author={Hui, Tianrui and Xun, Zizheng and Peng, Fengguang and Huang, Junshi and Wei, Xiaoming and Wei, Xiaolin and Dai, Jiao and Han, Jizhong and Liu, Si},
  booktitle={CVPR},
  pages={13630--13639},
  year={2023}
}

@article{cat++,
  title={RGBT Tracking via Challenge-Based Appearance Disentanglement and Interaction},
  author={Liu, Lei and Li, Chenglong and Xiao, Yun and Ruan, Rui and Fan, Minghao},
  journal={TIP},
  volume={33},
  pages={1753--1767},
  year={2024},
  publisher={IEEE}
}

@inproceedings{apfnet,
  title={Attribute-based progressive fusion network for rgbt tracking},
  author={Xiao, Yun and Yang, Mengmeng and Li, Chenglong and Liu, Lei and Tang, Jin},
  booktitle={AAAI},
  volume={36},
  number={3},
  pages={2831--2838},
  year={2022}
}

@article{stmt,
  title={Transformer RGBT Tracking With Spatio-Temporal Multimodal Tokens},
  author={Sun, Dengdi and Pan, Yajie and Lu, Andong and Li, Chenglong and Luo, Bin},
  journal={TCSVT},
  volume={34},
  number={11\_Part\_2},
  pages={12059--12072},
  year={2024},
  publisher={IEEE}
}

@article{mplt,
  title={RGB-T tracking via multi-modal mutual prompt learning},
  author={Luo, Yang and Guo, Xiqing and Feng, Hui and Ao, Lei},
  journal={arXiv preprint arXiv:2308.16386},
  year={2023}
}

@inproceedings{vipt,
  title={Visual prompt multi-modal tracking},
  author={Zhu, Jiawen and Lai, Simiao and Chen, Xin and Wang, Dong and Lu, Huchuan},
  booktitle={CVPR},
  pages={9516--9526},
  year={2023}
}

@inproceedings{bat,
  title={Bi-directional adapter for multimodal tracking},
  author={Cao, Bing and Guo, Junliang and Zhu, Pengfei and Hu, Qinghua},
  booktitle={AAAI},
  volume={38},
  number={2},
  pages={927--935},
  year={2024}
}

@inproceedings{tatrack,
  title={Temporal adaptive rgbt tracking with modality prompt},
  author={Wang, Hongyu and Liu, Xiaotao and Li, Yifan and Sun, Meng and Yuan, Dian and Liu, Jing},
  booktitle={AAAI},
  volume={38},
  number={6},
  pages={5436--5444},
  year={2024}
}

@inproceedings{sdstrack,
  title={Sdstrack: Self-distillation symmetric adapter learning for multi-modal visual object tracking},
  author={Hou, Xiaojun and Xing, Jiazheng and Qian, Yijie and Guo, Yaowei and Xin, Shuo and Chen, Junhao and Tang, Kai and Wang, Mengmeng and Jiang, Zhengkai and Liu, Liang and others},
  booktitle={CVPR},
  pages={26551--26561},
  year={2024}
}

@article{cfbt,
  title={Cross Fusion RGB-T Tracking with Bi-directional Adapter},
  author={Zeng, Zhirong and Liu, Xiaotao and Sun, Meng and Wang, Hongyu and Liu, Jing},
  journal={CoRR},
  year={2024}
}

@inproceedings{ckd,
  title={Breaking modality gap in RGBT tracking: Coupled knowledge distillation},
  author={Lu, Andong and Zhao, Jiacong and Li, Chenglong and Xiao, Yun and Luo, Bin},
  booktitle={ACM MM},
  pages={9291--9300},
  year={2024}
}

@inproceedings{sutrack,
  title={Sutrack: Towards simple and unified single object tracking},
  author={Chen, Xin and Kang, Ben and Geng, Wanting and Zhu, Jiawen and Liu, Yi and Wang, Dong and Lu, Huchuan},
  booktitle={AAAI},
  volume={39},
  number={2},
  pages={2239--2247},
  year={2025}
}

@inproceedings{sttrack,
  title={Exploiting multimodal spatial-temporal patterns for video object tracking},
  author={Hu, Xiantao and Tai, Ying and Zhao, Xu and Zhao, Chen and Zhang, Zhenyu and Li, Jun and Zhong, Bineng and Yang, Jian},
  booktitle={AAAI},
  volume={39},
  number={4},
  pages={3581--3589},
  year={2025}
}

@article{seqtrackv2,
  title={Unified Sequence-to-Sequence Learning for Single-and Multi-Modal Visual Object Tracking},
  author={Chen, Xin and Kang, Ben and Zhu, Jiawen and Wang, Dong and Peng, Houwen and Lu, Huchuan},
  journal={arXiv preprint arXiv:2304.14394},
  year={2023}
}

@inproceedings{ustrack,
  title={Unified Single-Stage Transformer Network for Efficient RGB-T Tracking},
  author={Xia, Jianqiang and Shi, Dianxi and Song, Ke and Song, Linna and Wang, Xiaolei and Jin, Songchang and Zhao, Chenran and Cheng, Yu and Jin, Lei and Zhu, Zheng and others},
  booktitle={IJCAI},
  year={2024}
}

@inproceedings{untrack,
  title={Single-model and any-modality for video object tracking},
  author={Wu, Zongwei and Zheng, Jilai and Ren, Xiangxuan and Vasluianu, Florin-Alexandru and Ma, Chao and Paudel, Danda Pani and Van Gool, Luc and Timofte, Radu},
  booktitle={CVPR},
  pages={19156--19166},
  year={2024}
}

@article{dmd,
  title={Dual-Level Modality De-Biasing for RGB-T Tracking},
  author={Hu, Yufan and Shao, Zekai and Fan, Bin and Liu, Hongmin},
  journal={TIP},
  year={2025},
  publisher={IEEE}
}

@inproceedings{pura,
  title={PURA: Parameter Update-Recovery Test-Time Adaption for RGB-T Tracking},
  author={Shao, Zekai and Hu, Yufan and Fan, Bin and Liu, Hongmin},
  booktitle={CVPR},
  pages={22089--22098},
  year={2025}
}

@article{lasher,
  title={LasHeR: A large-scale high-diversity benchmark for RGBT tracking},
  author={Li, Chenglong and Xue, Wanlin and Jia, Yaqing and Qu, Zhichen and Luo, Bin and Tang, Jin and Sun, Dengdi},
  journal={TIP},
  volume={31},
  pages={392--404},
  year={2021},
  publisher={IEEE}
}

@inproceedings{rgbt210,
  title={Weighted sparse representation regularized graph learning for RGB-T object tracking},
  author={Li, Chenglong and Zhao, Nan and Lu, Yijuan and Zhu, Chengli and Tang, Jin},
  booktitle={ACM MM},
  pages={1856--1864},
  year={2017}
}

@article{rgbt234,
  title={RGB-T object tracking: Benchmark and baseline},
  author={Li, Chenglong and Liang, Xinyan and Lu, Yijuan and Zhao, Nan and Tang, Jin},
  journal={PR},
  volume={96},
  pages={106977},
  year={2019},
  publisher={Elsevier}
}

@article{gtot,
  title={Learning collaborative sparse representation for grayscale-thermal tracking},
  author={Li, Chenglong and Cheng, Hui and Hu, Shiyi and Liu, Xiaobai and Tang, Jin and Lin, Liang},
  journal={TIP},
  volume={25},
  number={12},
  pages={5743--5756},
  year={2016},
  publisher={IEEE}
}

@inproceedings{lora,
  title={LoRA: Low-Rank Adaptation of Large Language Models},
  author={Hu, Edward J and Wallis, Phillip and Allen-Zhu, Zeyuan and Li, Yuanzhi and Wang, Shean and Wang, Lu and Chen, Weizhu and others},
  booktitle={ICLR},
  year={2022}
}

@article{hydralora,
  title={Hydralora: An asymmetric lora architecture for efficient fine-tuning},
  author={Tian, Chunlin and Shi, Zhan and Guo, Zhijiang and Li, Li and Xu, Cheng-Zhong},
  journal={NeurIPS},
  volume={37},
  pages={9565--9584},
  year={2024}
}

@inproceedings{adalora,
  title={Adaptive Budget Allocation for Parameter-Efficient Fine-Tuning},
  author={Zhang, Qingru and Chen, Minshuo and Bukharin, Alexander and He, Pengcheng and Cheng, Yu and Chen, Weizhu and Zhao, Tuo},
  booktitle={ICLR},
  year={2023},
  organization={Openreview}
}

@inproceedings{dora,
  title={Dora: Weight-decomposed low-rank adaptation},
  author={Liu, Shih-Yang and Wang, Chien-Yi and Yin, Hongxu and Molchanov, Pavlo and Wang, Yu-Chiang Frank and Cheng, Kwang-Ting and Chen, Min-Hung},
  booktitle={ICML},
  year={2024}
}

@article{mola,
  title={Higher layers need more lora experts},
  author={Gao, Chongyang and Chen, Kezhen and Rao, Jinmeng and Sun, Baochen and Liu, Ruibo and Peng, Daiyi and Zhang, Yawen and Guo, Xiaoyuan and Yang, Jie and Subrahmanian, VS},
  journal={arXiv preprint arXiv:2402.08562},
  year={2024}
}

@article{mixlora,
  title={MixLoRA: Enhancing Large Language Models Fine-Tuning with LoRA based Mixture of Experts},
  author={Li, Dengchun and Ma, Yingzi and Wang, Naizheng and Cheng, Zhiyuan and Duan, Lei and Zuo, Jie and Yang, Cal and Tang, Mingjie},
  journal={CoRR},
  year={2024}
}

@article{ia3,
  title={Few-shot parameter-efficient fine-tuning is better and cheaper than in-context learning},
  author={Liu, Haokun and Tam, Derek and Muqeeth, Mohammed and Mohta, Jay and Huang, Tenghao and Bansal, Mohit and Raffel, Colin A},
  journal={NeurIPS},
  volume={35},
  pages={1950--1965},
  year={2022}
}

@inproceedings{vpt,
  title={Visual prompt tuning},
  author={Jia, Menglin and Tang, Luming and Chen, Bor-Chun and Cardie, Claire and Belongie, Serge and Hariharan, Bharath and Lim, Ser-Nam},
  booktitle={ECCV},
  pages={709--727},
  year={2022},
  organization={Springer}
}

@inproceedings{adapter,
  title={Parameter-efficient transfer learning for NLP},
  author={Houlsby, Neil and Giurgiu, Andrei and Jastrzebski, Stanislaw and Morrone, Bruna and De Laroussilhe, Quentin and Gesmundo, Andrea and Attariyan, Mona and Gelly, Sylvain},
  booktitle={ICML},
  pages={2790--2799},
  year={2019},
  organization={PMLR}
}

@inproceedings{prompt-tuning,
  title={The Power of Scale for Parameter-Efficient Prompt Tuning},
  author={Lester, Brian and Al-Rfou, Rami and Constant, Noah},
  booktitle={EMNLP},
  pages={3045--3059},
  year={2021}
}

@inproceedings{prefix-tuning,
  title={Prefix-Tuning: Optimizing Continuous Prompts for Generation},
  author={Li, Xiang Lisa and Liang, Percy},
  booktitle={ACL},
  pages={4582--4597},
  year={2021}
}

@incollection{convpass,
  title={Convolutional bypasses are better vision transformer adapters},
  author={Jie, Shibo and Deng, Zhi-Hong and Chen, Shixuan and Jin, Zhijuan},
  booktitle={ECAI},
  pages={202--209},
  year={2024},
  publisher={IOS Press}
}

@inproceedings{giou,
  title={Generalized intersection over union: A metric and a loss for bounding box regression},
  author={Rezatofighi, Hamid and Tsoi, Nathan and Gwak, JunYoung and Sadeghian, Amir and Reid, Ian and Savarese, Silvio},
  booktitle={CVPR},
  pages={658--666},
  year={2019}
}

@article{dinov2,
  title={DINOv2: Learning Robust Visual Features without Supervision},
  author={Oquab, Maxime and Darcet, Timoth{\'e}e and Moutakanni, Th{\'e}o and Vo, Huy V and Szafraniec, Marc and Khalidov, Vasil and Fernandez, Pierre and HAZIZA, Daniel and Massa, Francisco and El-Nouby, Alaaeldin and others},
  journal={TMLR},
  year={2024}
}

@article{c-kmeans,
  title={Constrained K-Means Clustering},
  author={Bradley, PS and Bennett, KP and Demiriz, A},
  year={2000}
}

\end{document}